\documentclass[11pt]{article}

\date{}

\usepackage[utf8]{inputenc} 
\usepackage[T1]{fontenc}    
\usepackage{hyperref}       
\usepackage{url}            
\usepackage{booktabs}       
\usepackage{amsfonts}       
\usepackage{nicefrac}       
\usepackage{microtype}      
\usepackage{xspace}
\usepackage{natbib}

\usepackage{comment}
\usepackage{amsmath}
\usepackage{amsthm}
\usepackage{graphicx}
\usepackage{rotating}
\usepackage{float}
\usepackage{breakcites}
\usepackage{microtype}

\usepackage{amssymb}
\usepackage{autobreak}
\usepackage{mathtools}
\usepackage{xcolor}
\usepackage{bbm}
\usepackage{algorithm}
\usepackage{algorithmic}

\makeatletter

\newcommand{\GTRPO}{\textsc{\small{GTRPO}}\xspace}
\newcommand{\RL}{\textsc{\small{RL}}\xspace}
\newcommand{\DRL}{\textsc{\small{DRL}}\xspace}
\newcommand{\PG}{\textsc{\small{PG}}\xspace}

\newcommand{\GPPO}{\textsc{\small{GPPO}}\xspace}
\newcommand{\POMDP}{\textsc{\small{POMDP}}\xspace}
\newcommand{\MDP}{\textsc{\small{MDP}}\xspace}
\newcommand{\TRPO}{\textsc{\small{TRPO}}\xspace}
\newcommand{\PPO}{\textsc{\small{PPO}}\xspace}

\renewcommand{\Re}{\mathbb{R}}
\newcommand{\calP}{\mathcal P}
\newcommand{\PP}{\mathbb P}
\newcommand{\F}{\mathcal F}

\newcommand{\A}{\mathcal A}

\newcommand{\X}{\mathcal X}
\newcommand{\T}{\mathcal T}
\newcommand{\KL}{\mathcal D_{KL}}
\newcommand{\wKL}{\wh{\mathcal{D}}_{KL}}
\newcommand{\D}{\mathcal D}
\newcommand{\Gr}{\nabla_{\theta}}

\newcommand{\Grrho}{\nabla \rho}

\newcommand{\Y}{\mathcal Y}

\newcommand{\E}{\mathbb E}

\newcommand{\wt}{\widetilde}
\newcommand{\wh}{\widehat}
\newcommand{\wb}{\overline}

\newtheorem{lemma}{Lemma}[section]
\newtheorem{assumption}{Assumption}[section]
\newtheorem{theorem*}{Theorem}[section]
\newtheorem{corollary}{Corollary}[section]

\newtheorem{definition}{Definition}[section]
\newtheorem{theorem}{Theorem}[section]
\newtheorem{remark}{Remark}[section]

\DeclareMathOperator*{\argmax}{arg\,max}

\newcommand*{\lem}{Lem.~}

\newcommand*{\thm}{Thm.~}

\title{Policy Gradient in Partially Observable Environments:\\
Approximation and Convergence}

\author{%
 Kamyar Azizzadenesheli, Yisong Yue, Anima Anandkumar\\
Department of Computing and Mathematical Sciences\\
  California Institute of Technology, Pasadena\\
  \texttt{\{kazizzad,yyue,anima\}@caltech.edu} \\
}

\begin{document}

\maketitle

\begin{abstract}%
Policy gradient is a generic and flexible reinforcement learning approach that generally enjoys simplicity in analysis, implementation, and deployment. In the last few decades, this approach has been extensively advanced for fully observable environments. In this paper, we generalize a variety of these advances to partially observable settings, and similar to the fully observable case, we keep our focus on the class of Markovian policies. We propose a series of technical tools, including a novel notion of advantage function, to develop policy gradient algorithms and study their convergence properties in such environments. Deploying these tools, we generalize a variety of existing theoretical guarantees, such as policy gradient and convergence theorems, to partially observable domains, those which also could be carried to more settings of interest. This study also sheds light on the understanding of policy gradient approaches in real-world applications which tend to be partially observable.


\end{abstract}

\section{Introduction}
\begin{sloppypar}
Reinforcement learning (\RL) is the study of sequential decision making under uncertainty with a vast application in real-world problems such as robotics, ad-allocation, recommendation, and autonomous vehicles systems. In \RL, a decision-making agent interacts with its surrounding environment, collects experiences throughout the interaction, and develops its understanding of the environment. The agent exploits this information to improve its behavior and strategies for further exploration.\footnote{Note that the described setting is one the primary setting of study in \RL} One of the main problems in \RL is the design of efficient algorithms that improve the agent's performance in high dimensional environments and maximize a notion of reward.
 
Recent advances in \RL, particularly in deep \RL (\DRL), have shown major enhances in high-dimensional \RL problems. Among the most popularized ones, \citet{abbasi2011regret} addresses control in linear systems, \citet{mnih2015human} proposes Deep-Q networks and tackles arcade games~\citep{bellemare2013arcade}, \citet{silver2016mastering} uses upper confidence bound tree search~\citep{kocsis2006bandit} and tackles board games, and~\citet{schulman2015trust} employs policy gradient (\PG)~\citep{Aleksandrov68} and addresses continuous control problems, especially MuJoCo~\citep{todorov2012mujoco} environments. 
\end{sloppypar}

This paper is concerned with \PG methods, which are among the prominent methods in high-dimensional \RL. \PG methods are gradient optimization-based, and mainly approach \RL  as stochastic optimization problems. These methods usually deploy Monte Carlo sampling to estimate the gradient updates, resulting in high variance gradient estimations~\citep{Rubinstein69}. Recent works study Markov decision processes (\MDP{}) and exploit the \MDP structure to mitigate this high variance issue~\citep{sutton2000policy,kakade2002approximately,kakade2002natural,schulman2015trust,lillicrap2015continuous}. These works employ value-based methods, and provide low variance \PG methods, mainly in infinite horizon discounted reward \MDP{}s, and pave the road to guarantee monotonic improvements in the performance~\citep{kakade2002approximately}. However, they do not offer optimality guarantees.

Understanding the convergence characteristics of \PG methods, along with their global optimality properties, and relationship with problem-specific dependencies, are essential for effective algorithm design in real-world applications. Recently, a study by \citet{fazel2018global} advances the understanding of the optimization landscape of the infinite horizon linear quadratic problems and proposes a set of algorithms which are guaranteed to converge to the optimal control under a coverage assumption. Further encouraging studies provide convergence analyses of \PG related methods in \MDP{}s mainly using fixed-point convergence arguments~\citep{liu2015finite,bhandari2019global,agarwal2019optimality,wang2019neural,abbasi2019politex}. These analyses advance the understating of \PG methods related in infinite-horizon \MDP{}s.

These recent developments  of \PG methods, both theoretical and empirical,  are confined to fully observable setting \MDP{}s.  However, in many real-world \RL applications, e.g., robotic, drone delivery, and navigation,  the underlying states of the environments are hidden, i.e., they are more appropriately modeled as partially observable \MDP{}s. 
In \POMDP{}s, the observation process need not be Markovian, and only incomplete information of the underlying state of the environment is accessible to the decision-makers.

\textbf{In this work}, we theoretically study \PG methods in episodic \POMDP{}s, and analyze their convergence properties and optimality guarantees. We establish these results for both discounted and undiscounted reward scenarios. We also develop a series of \PG algorithms, agnostic to the underlying environment model, fully (\MDP{}s), or partially observable (\POMDP{}s).\footnote{Note that, since the class of \MDP{}s is a subset to the class of \POMDP{}s, under a proper choice of policy class, designing algorithms for \POMDP{}s suffices to have agnostic algorithms to the underlying model.} To design algorithms that are agnostic to the choice of the underlying model, we adopt the class of Markovian policies~\citep{puterman2014markov}, which, under certain regularity conditions, are optimal for \MDP{}s. We propose a series of technical tools, including a novel notion of advantage function along with value and Q functions for \POMDP{}s, that we employ to analyze \PG methods.

We analyze and generalize a range of \MDP{} methods to the \POMDP{} setting, such as trust region policy optimization (\TRPO) \citep{schulman2015trust}, and proximal policy optimization (\PPO) \citep{schulman2017proximal}.  The resulting generalized \TRPO (\GTRPO) and generalized \PPO (\GPPO) methods are among the few for \POMDP{}s that are computationally feasible. We conclude our study by showing how the tools developed in this work make it feasible to generalize a variety of existing \MDP based advanced algorithms to \POMDP{}s. This study sheds light on the understanding of \PG approaches in real-world applications that are inherently partially observable.


\paragraph{Extended motivation, problem setting, and contribution.} ~ \\

In the following, we provide a detailed introduction of the setting of study in this paper, explain hardness results, and conclude with a detailed explanation of our contributions.
Table~\ref{table:category} categorizes the majority of the \RL paradigm concerning their observability, policy class, horizon, and discounting factor.\footnote{Note that this categorization is a coarse summary and does not capture all \RL paradigms.}

\newcommand\dunderline[3][-1pt]{{%
  \setbox0=\hbox{#3}
  \ooalign{\copy0\cr\rule[\dimexpr#1-#2\relax]{\wd0}{#2}}}}

\newcommand{\rul}[1]{\textcolor{red}{\dunderline{1pt}{\textcolor{black}{#1}}}}
\newcommand{\bul}[1]{\textcolor{blue}{\dunderline{3pt}{\textcolor{black}{#1}}}}

\begin{table}[t!]
  \centering
  \caption{ While the majority of the prior works study \MDP{}s and Markovian policies under discounted reward and infinite horizon considerations, we focus our study on episodic problems, no matter \MDP or \POMDP, Markovian or non-Markovian, and discounted or undiscounted.}
  \footnotesize
  \begin{tabular}{c|c|c|c}
    \toprule
 Observability  & Policy Class &  Discounting & Horizon \\
  \midrule
MDP & Markovian& Discounted & Infinite \\
POMDP & non-Markovian& Undiscounted  & Episodic \\
    \bottomrule
  \end{tabular}
  \label{table:category}
\end{table}


\textit{Episodic and infinite horizon}: Many contemporary applications of \RL are cast as episodic and/or finite horizon problems. Episodic \RL refers to settings where learning happens iteratively over episodes, with each episode comprising running the current policy from a (often fixed) distribution of initial states, collecting data, and updating the policy.  Episodic \RL is thus appropriate for any sequential decision making tasks that are run in a repeated fashion.  Infinite horizon refers to casting the problem as running each episode over for infinite time steps. Episodic, finite-horizon test-bed environments, such as MuJoCo, have played a major role in the recent empirical successes of algorithm design in \RL, despite many of those algorithms being designed for the infinite horizon setting.
While the study of infinite horizon environments is an essential topic of research, we dedicate this paper to episodic environments.

\textit{Discounted and undiscounted reward}:
Another distinction, is whether to use discounted or undiscounted cumulative rewards. 
In the discounted cumulative reward settings, rewards that are received earlier in time are more favorable to those received later in time. In this setting, discounted cumulative rewards are accumulations of rewards that are (typically exponentially) discounted through time. In contrast, in the undiscounted cumulative reward settings, \RL agents are indifferent to when a unit of reward is received, as long as it is received.  In repetitive related tasks with a clear completion goal (e.g., cooking), we might be interested in the episodic undiscounted reward with fixed horizons. In such tasks, we might be concerned with task accomplishment in a fixed period of time. In contrast, long-horizon tasks with accumulated goals, such as vacuuming, we might rather have a given physical area to be cleaned earlier than later without specifically specifying a task horizon. Discounted rewards also favor completing tasks early rather than later, owing to putting lower weight on rewards in future time steps. 
While the previous theoretical development of \PG methods ~\citep{schulman2015trust,lillicrap2015continuous} are mainly dedicated to the discounted reward setting (also infinite horizon), we extend our study to both the discounted and undiscounted reward settings.



\textit{Fully and partially observable}:
While \PG can in principle be applied very generically, as mentioned above, previous research on advanced \PG methods has larged focused on (fully observable) \MDP{}s. Many real-world learning and decision-making tasks, however, are inherently partially observable where an incomplete representation of the state of the environment is accessible. Owing to the inability to directly observe the (latent) underlying states, learning in \POMDP environments poses significant challenges and also adds computational complexity to the resulting policy optimization problem. Moreover, it is known that applying \MDP based methods on \POMDP{}s can result in policies with arbitrarily bad expected returns~\citep{williams1999experimental,azizzadenesheli2017experimental}. In this work, we study environment agnostic \PG methods and provide convergence as well as monotonic improvement guarantees for both \MDP{}s and \POMDP{}s.

\textit{Markovian and non-Markovian policies}:
Under a few conditions~\citep{puterman2014markov}, for any given \MDP{}, there exists an optimal policy which is deterministic and Markovian (that maps the agent's immediate or current observation of the environment to actions). In contrast, when dealing with \POMDP{}s, we might be interested in the class of non-Markovian and history-dependent policies (that map the state-action trajectory history to distributions over actions). However, tackling non-Markovian and history-dependent policies can be computationally undecidable~\citep{madani1999undecidability} for the infinite horizon, or PSPACE-Complete~\citep{papadimitriou1987complexity} in the finite-horizon \POMDP{}s. 
To avoid undesirable computational burdens, we focus on Markovian policies.

Indeed, many prior works study Markovian policies for \POMDP{}s~\citep{baxter2001infinite,littman1994memoryless,Baxter_2000,azizzadenesheli2016reinforcement} where, in general, the optimal policies are stochastic~\citep{littman1994memoryless,singh1994learning,montufar2015geometry}. Acknowledging the computation complexity of Markovian policies~\citep{vlassis2012computational}, this line of work highlights the broad interest, importance, and applicability of Markovian policies. 

We are further interested in algorithms that are agnostic to the underlying model, as restricting to such algorithm designs can also avoid certain undesirable computational burdens.
We do so by choosing Markovian policies, which are generally optimal for \MDP{}s. This choice of the policy class allows us to use that same class of functions to represent policies in \POMDP{}s. We can thus carry over \PG results from  \MDP{}s directly to their \POMDP{} counterparts. 
An additional practical advantage is that it is straightforward to re-purpose well-developed software implementations of \MDP{}-based \PG algorithms for the \POMDP{} setting.

One could alternatively consider history-dependent policies, which is a richer function class than Markovian policies.  However, utilizing history-dependent policies requires turning a given \POMDP problem to a potentially non-stationary \MDP with states as concatenations of the historical data.  
As such, a thorough theoretical treatment of history-dependent policies for \POMDP{}s would likely require leveraging theoretical analyses for non-stationary \MDP{}s.



\textit{Equivalence policy classes and the role of observation boundary}:
The expressiveness of general policy classes is mainly entangled with the definition of the observation. Under some regularity conditions, for any given class of history-dependent policies on a \POMDP, there exists a class of Markovian policies on a new \POMDP such that: $(1)$ the observations of the new \POMDP are the (typically) discrete concatenations of the historical data in the former \POMDP; and $(2)$ the two policy classes are equivalent, i.e., the respective policies result in the same behavior (e.g., action sequence). In other words, instead of making the policies non-Markovian and history-dependent, we can keep them Markovian and instead enrich the observation space. Similarly, for any limited-memory policy class (depending on a fixed window of history instead of the whole history, e.g., the policies in \MDP{}s of order more than one), there is an enrichment of the observation that results in an equivalent class of Markovian policies. This observation-enrichment viewpoint is known as the emergentism approach.\footnote{In contrast, the reductionism approach argues for the opposite viewpoint, and views the class of Markovian policies is a subset of the class of non-Markovian and history-dependent policies, simply by ignoring the historical data and considering just the immediate/current observation for decision making. While both viewpoints are valid, they differ in exercising the definition of observations.} 
In essence, these distinctions boil down to: ``What a Markovian policy is Markovian with respect to''. 


The above reasoning implies that, by considering the class of Markovian policies in episodic \POMDP{}s, we often will not restrict the generality of the results in this paper. Therefore, despite focusing on Markovian policies, our results also hold for both classes of limited-memory as well as history-dependent policies through representing histories as observations. \\


\textbf{Detailed Summary of Contributions.} In this paper, we study Markovian policies in episodic \POMDP{}s and \MDP{}s, with both discounted and discounted rewards.

$(i)$ We state the general \PG theorems in the \POMDP{} framework. These theorems are based on are well-known theorems that make no specific modeling assumption on the underlying environments. We extend the value-based \PG theorems on \MDP{}, such as deterministic \PG\citep{silver2014deterministic}, to \POMDP{}s.  While the extensions are straightforward, we provide them for clarity and completeness.

$(ii)$ We define a novel notion of advantage function for \POMDP{}s and advance a series of \MDP{}-based theoretical results to \POMDP{}s. In \MDP{}s, the advantage functions are well defined, and they are functions of one event (i.e., one state-action pair). However, the states in \POMDP{}s are partially observed, and the classical definition of advantage functions does not directly carry to partially observable cases. For \POMDP{}s, our advantage function depends on three consecutive events instead of just one. Utilizing this definition, we generalize the policy improvement guarantees in \MDP{}s \citep{kakade2002approximately,schulman2015trust} to \POMDP{}s. Note that conditioning on three consecutive events for learning in \POMDP{}s matches the results in~\citet{azizzadenesheli2016reinforcement}, which indicate that the statistics of three consecutive events are sufficient to learn \POMDP dynamics and guarantee order-optimal sample complexity properties of Markovian policies. 

$(iii)$
We study how to extend natural policy gradient methods for \MDP{}s to  \POMDP{}s.  This derivation leverages our novel notion of advantage function.

\begin{sloppypar}

$(iv)$ We propose generalized trust region policy optimization (\GTRPO) \PG methods, as a generalization of ~\citet{kakade2002approximately,schulman2017proximal} to the general classes of episodic  \MDP{}s and   \POMDP{}s under both discounted and undiscounted reward settings. \GTRPO utilize the property of newly developed advantage function, and computes a low variance estimation of the \PG updates. We then construct a trust region for the policy search step in \GTRPO and show that the policy updates of \GTRPO are guaranteed to monotonically improve the expected return. \GTRPO is among the few methods for \POMDP{}s that are computationally feasible. 

$(v)$ We propose generalized proximal policy optimization (\GPPO), a generalization of the \PPO algorithm~\citep{schulman2017proximal} to the general classes of episodic  \MDP{}s and   \POMDP{} under both discounted and undiscounted reward settings. \GPPO is an extension to \GTRPO, as \PPO is an extension of \TRPO, which is known to be computation more efficient than its predecessor.
\end{sloppypar}

We conclude the paper by stating that the developed machinery in this paper, can be used to extend a variety of existing \MDP{}-based methods directly to their general \POMDP cases.






\section{Preliminaries}


\begin{figure}[t]
\vspace{-.1cm}
\begin{center}
\includegraphics[scale=0.25]{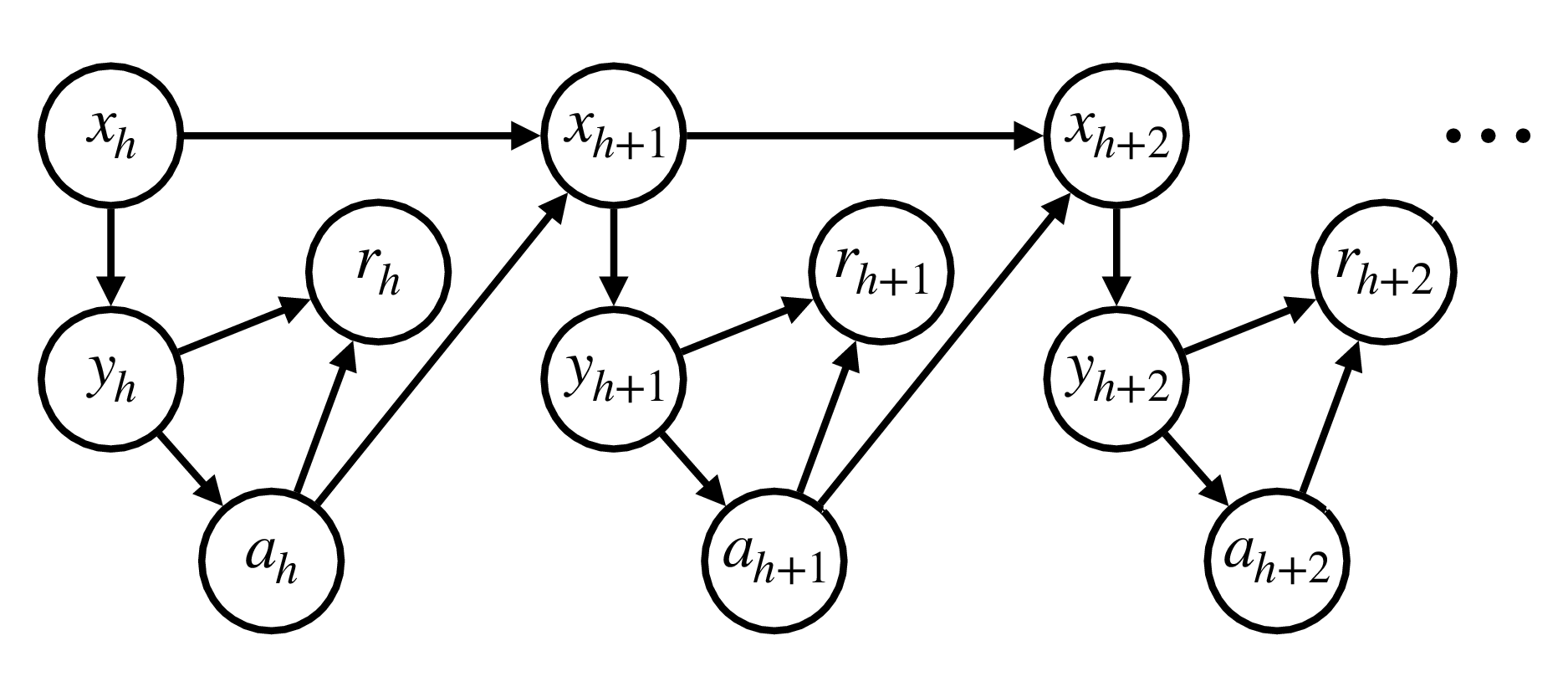}
\end{center}
\caption{\POMDP under a Markovian policy}
\label{fig:pds2}
\end{figure}

An episodic stochastic \POMDP, $M$, is a tuple $\langle \X, \A, \Y, P_1,T, R, O,\gamma, x_{Terminal}\rangle$ with a set of latent states $\X$, observations $\Y$, actions $\A$, and their elements denoted as $x\in\X$, $y\in\Y$, and $a\in\A$, accompanied with a discount factor $0\leq\gamma\leq1$. Under reasonable assumptions,\footnote{We assume that the quantities of interests are measurable, and Lebesgue-Stieljes integrable with respect to their corresponding kernels on their respective Borel measurable spaces, among other regularity assumptions, such as compactness of the action sets, complete and separability of $\X,\Y$ spaces, as well as continuity of value related functions. For simplicity, wherever possible, we also omit the proper measure theoretic definition of the stochastic processes as well as the generality of functions classes, unless necessary.} let $P_1(\cdot): \X\rightarrow [0,\infty]$ represent the probability distribution of the initial latent states, $T(\cdot|\cdot,\cdot):\X\times \X\times\A\rightarrow [0,\infty]$ denotes the transition kernel on the latent states, $O(\cdot|\cdot):\Y\times \X\rightarrow [0,\infty]$ the emission kernel, $R(\cdot|\cdot,\cdot):\Re\times \Y\times \A\rightarrow [0,\infty]$ the reward kernel.\footnote{In an alternative definition, the reward kernel follows the latent states, i.e., $R(\cdot|\cdot,\cdot):\Re\times \X\times \A\rightarrow [0,\infty]$.} Let the kernel map $\pi(\cdot|\cdot):\A\times \Y\rightarrow [0,\infty]$ denote a Markovian policy.
%
%
For simplicity in the notation, we encode the time step $h$ into the state, observation, and action representations, $x$, $y$, $a$. 
Consider the following event in the stochastic process ruled by $M$ and a policy $\pi$, in short, a trajectory:
\begin{align*}
\tau:=\lbrace (x_1,y_1,a_1,r_1),(x_2,y_2,a_2,r_2),\ldots, (x_{|\tau|},y_{|\tau|},a_{|\tau|},r_{|\tau|}),x_{|\tau|+1},y_{|\tau|+1}\rbrace,
\end{align*}
where $|\tau|$ is the random length of the trajectory. 
In episodic setting, a policy $\pi$ interacts with the environment as follows (extended definitions follow the description of the interaction protocol):
    The policy interacts with the environment in episodes:
    \begin{enumerate}
        \item The policy starts at an initial state sampled from the initial state distribution, $x_1\sim P_1$.
        \item For each time step $h\geq1$, the policy receives observations $y_h \sim O(\cdot|x_h)$.
        \item If $x_{h} = x_{Terminal}$, then $|\tau|$ is $h-1$, and end the episode.
        \item The policy then takes action $a_h \sim \pi(\cdot|y_h)$, and receives reward $r_h \sim R(\cdot|y_h,a_h)$.
        \item The environment transitions to a new state $x_{h+1} \sim T(\cdot|x_h,a_h)$. 
        \item $h \leftarrow h+1$, repeat from Step (a).
    \end{enumerate}

Since $M$ is an episodic environment, $\PP(|\tau|< \infty) = 1$. 
%
As the definition of episodic environments indicates, $x_{Terminal}$ is reachable in finite random time $|\tau|+1$, i.e., $|\tau|+1<\infty$ almost surely. Correspondingly, define $y_{Terminal}:=y_{|\tau|+1}$ as the terminal observation. 
For any given pairs of $(y,a)\in\Y\times\A$, $\wb{R}(y,a) := \mathbb{E}[R(y,a)]$ denotes a version of conditional expected of reward where the expectation is with respect to the randomness of reward given $y$ and $a$. We assume $\wb{R}(\cdot,\cdot)$ is a finite function. Fig.~\ref{fig:pds2} depicts a Markovian policy acting on a \POMDP{}. 

We consider a set $\calP$ of parameterized Markovian policies $\pi_{\theta}$ where $\theta\in\Theta$, with $\Theta$ an appropriate compact euclidean space, i.e., a compact subset of finite Cartesian product of space of reals.\footnote{For simplicity in the derivations, we assume that the value related functions in this paper are continuous in $\Theta$.} Each policy $\pi_{\theta}$ in $\calP$ is also assumed to be continuously differentiable with respect to parameter $\theta\in\Theta$, with finite gradients (space of $W^{1,\infty}$). For any given parameter $\theta$, the construction of $M$ results in a set of trajectories, such that $\tau\in\Upsilon$ and $(\Upsilon,\F,\PP_\theta)$ is a complete measure space with $\F$, the corresponding $\sigma$-algebra. For simplicity, let $\forall \tau\in\Upsilon, ~f(\tau;\theta)$ denote the probability distribution of trajectories under policy $\pi_{\theta}$, i.e.,
\begin{align*}
f(\tau;\theta)d\tau:=d\PP_\theta=P_1(x_1)\prod_{h=1}^{|\tau|}O(y_h|x_h)\pi_\theta(a_h|y_h)R(r_h|x_h,a_h)T(x_{h+1}|x_{h},a_{h})d\tau.
\end{align*}


For a trajectory $\tau$, a random variable $R(\tau)$ represent the cumulative $\gamma$-discounted rewards in $\tau$, i.e., $R(\tau)=\sum_{h=1}^{|\tau|}\gamma^{h-1}r_h$. Therefore, the unnormalized expected cumulative return of $\pi_{\theta}$ is:
\begin{align}\label{eq:eta}
\eta(\theta):=\mathbb{E}_{\tau|\theta}[R(\tau)]=\int_{\tau} f(\tau;\theta) R(\tau)d\tau \longrightarrow
\theta^*\in\arg\min_{\theta\in\Theta}\eta(\theta).
\end{align}
We use $\mathbb{E}_{\tau|\theta}$, $\mathbb{E}_{\tau\sim f(\cdot;\theta)}$, $\mathbb{E}_{\pi_\theta}$, interchangeably for  expectation with respect to the $d\PP_\theta$. Let $\theta^*$ denote a parameter that maximizes $\eta(\theta)$, i.e., $\theta^*\in\arg\max_{\theta\in\Theta}\eta(\theta)$, 
and $\pi^*=\pi(\theta^*)$ an optimal policy. The optimization problem to find an optimal policy in the policy class, in general, is non-convex and hill climbing methods such as gradient ascent may not necessary converge to the optimal behavior.

\section{Policy Gradient}
\PG is a generic \RL approach which directly optimizes for policies without requiring explicit construction of the environment model or the value functions.  Owing to its simplicity, PG is used in a variety of practical and theoretical setting. Theoretical analyses of this approach, known as \PG theorems, indicate that there is no need to have an explicit knowledge of the environment model, $M$, to compute the gradient of $\eta(\theta)$,~Eq.~\ref{eq:eta}, with respect to $\theta$ as long as we can sample returns from the environment. These theorems elaborate that a Monte Carlo sampling of returns suffices for the gradient computation \citep{Rubinstein69,williams1992simple,baxter2001infinite}.\footnote{In this paper we employ the measure-theoretic definition of the gradient, and consider large enough parameters space to avoid boundary related discussions.} In this section, we re-derive these theorems, but this time explicitly for \POMDP{}s under Markovian policies using two approaches: \textit{i}: through construction of the score functions, \textit{ii}: through importance sampling which is based on the change of measures and score functions. We emphasize on the importance sampling approach since it is more relevant to the later parts of this paper. 

In the Subsections \ref{sub:scorefunction},~\ref{sub:importanceweighting}, and~\ref{sub:Value-basedPG}, we extend the theoretical analyses of standard \PG theorems to the partially observed settings.  These extensions are straightforward, and present them for completeness to set up our main theoretical contributions in later Sections.

\subsection{Score Function}\label{sub:scorefunction}
As mentioned before, the gradient of the expected return can be estimated without explicit knowledge of the model dynamics and is agnostic to the realization of the underlying model. We re-establish this statement for \POMDP{} using score functions. Employing dominated convergence theorem, and the definition of $\eta(\theta)$ in Eq.~\ref{eq:eta} we have,

\begin{align*}
\nabla_{\theta}\eta(\theta)&=\nabla_{\theta}\int_{\tau} f(\tau;\theta) R(\tau)d\tau\\
&=\int_{\tau} \nabla_{\theta}f(\tau;\theta) R(\tau)d\tau\\
&=\int_{\tau} f(\tau;\theta)\nabla_{\theta} \log(f(\tau;\theta)) R(\tau)d\tau,
\end{align*}
where for a trajectory $\tau$, the score function is the gradient of the $\log(f(\tau;\theta))$ with respect to $\theta$, i.e.,
\begin{align*}
\nabla_{\theta} \log(f(\tau;\theta))&=\nabla_{\theta} \log\left( P_1(x_1)\Pi_{h=1}^{|\tau|}O(y_h|x_h)R(r_h|x_h,a_h)T(x_{h+1}|x_{h},a_{h})\right)+\nabla_{\theta}\log \left(\Pi_{h=1}^{|\tau|}\pi_\theta(a_h|y_h)\right)
\end{align*}
Since the first part of the right hand side of the above equation is independent of $\theta$, its derivative with respect to $\theta$ is zero. Therefore we have,
\begin{align*}
\nabla_{\theta}\eta(\theta)=\int_{\tau\in\Upsilon} f(\tau;\theta)\Gr\log \left(\Pi_{h=1}^{|\tau|}\pi_\theta(a_h|y_h)\right)R(\tau)d\tau.
\end{align*}

Following $\pi_\theta$, generate $m$ trajectories $\lbrace \tau^1,\ldots,\tau^m\rbrace$ with elements $(x^t_h,y^t_h,a^t_h,r^t_h)$, and terminals $\forall h\in\lbrace 1,\ldots,|\tau^{t}|\rbrace$ and $\forall t\in\lbrace 1,\ldots,m\rbrace$. Deploying the Monte Carlo sampling approximation of the integration, the empirical mean of the return is $\wh\eta(\theta)=\frac{1}{m}\sum_{t=1}^{m}R(\tau^t)$, and the estimation of the gradient is as follows,
\begin{align}
\nabla_{\theta}\wh\eta(\theta)=\frac{1}{m}\sum_{t=1}^{m}\nabla_{\theta}\log \left(\Pi_{h=1}^{|\tau^t|}\pi_\theta(a^t_h|y^t_h)\right)R(\tau^t).
\end{align}
This estimation is an unbiased and consistent estimation of the gradient and does not require the knowledge of the underlying dynamic except through samples of cumulative reward $R(\tau)$. 

%

\subsection{Importance Weighting}\label{sub:importanceweighting}
We derive the policy gradient theorem using techniques following the change of measures. We estimate $\eta(\theta')$, the expected return of policy $\pi_\theta'$ using trajectories induced by $\pi_\theta$ assuming the absolute continuity of imposed probably measure by $\theta'$ with respect to $\theta$'s. Again, using the definition of $\eta(\theta)$ in Eq.~\ref{eq:eta} we have:
\begin{align}\label{eq:IS_eta}
\eta(\theta')=\mathbb{E}_{\tau|\theta'}\left[R(\tau)\right]&=\int_{\tau\in\Upsilon} f(\tau;\theta')R(\tau)d\tau\nonumber\\
&=\int_{\tau\in\Upsilon} f(\tau;\theta)\left(\frac{f(\tau;\theta')}{f(\tau;\theta)}R(\tau)\right)d\tau=\mathbb{E}_{\tau|\theta}\left[\frac{f(\tau;\theta')}{f(\tau;\theta)}R(\tau)\right].
\end{align}
Again, using dominated convergence theorem, we compute the gradient of $\eta(\theta')$ with respect to $\theta'$:
\begin{align*}
\nabla_{\theta'}\eta(\theta') 
=\mathbb{E}_{\tau|\theta}\left[\frac{\nabla_{\theta'}f(\tau;\theta')}{f(\tau;\theta)}R(\tau)\right]
=\mathbb{E}_{\tau|\theta}\left[\frac{f(\tau;\theta')}{f(\tau;\theta)}\nabla_{\theta'}\log(f(\tau;\theta'))R(\tau)\right].
\end{align*}
If we take a limit $\theta'\rightarrow \theta$, and evaluate this gradient at $\theta'=\theta$, we have;
\begin{align}\label{eq:grad}
\nabla_{\theta'}\eta(\theta')\mid_{\theta'= ~\theta}=
\mathbb{E}_{\tau|\theta}\left[\nabla_{\theta}\log(f(\tau;\theta))R(\tau)\right].
\end{align}
Following a similar argument as in the Subsection~\ref{sub:scorefunction}, for a trajectory $\tau$ we have $\nabla_{\theta} \log(f(\tau;\theta))=\nabla_{\theta}\log \left(\Pi_{h=1}^{|\tau|}\pi_\theta(a_h|y_h)\right)$. Given $m$ trajectories $\lbrace \tau^1,\ldots,\tau^m\rbrace$, with elements $(x^t_h,y^t_h,a^t_h,r^t_h)$, and and terminals, $\forall h\in\lbrace 1,\ldots,|\tau^{t}|\rbrace$ and $\forall t\in\lbrace 1,\ldots,m\rbrace$ generated following $\pi_{\theta}$, we estimate the gradient $\nabla_{\theta}\eta(\theta)$ as follows, which is same expression as we derived using score function,
\begin{align}\label{eq:emp_eta_trajectory1}
\nabla_{\theta}\wh\eta(\theta)=\frac{1}{m}\sum_{t=1}^{m}\nabla_{\theta}\log \left(\Pi_{h=1}^{|\tau^t|}\pi_\theta(a^t_h|y^t_h)\right)R(\tau^t).
\end{align}
The gradient estimation using plain Monte Carlo samples incorporates samples of $R(\tau^t)$ which a long sum of random reward. Consequently, this estimation may result in a high variance estimation and, in practice, might not be an approach of interest. As it is evident, the \PG theorem is too powerful and, in expectation, is almost independent of the underlying stochastic process. In the next section, we describe how one can exploit the structure in the underlying \POMDP{} environment to develop a \PG method that mitigates the high variance shortcoming of direct Monte Carlo sampling.

\subsection{Value-base Policy Gradient Theorem}\label{sub:Value-basedPG}
To mitigate the effect of high variance gradient estimation, one might be interested in exploiting the environment structure and deploy the value-based methods. Let $\tau_{h'..h}$ denote the events in $\tau$ from the time step $h'$ up to the time step $h$, including the event at $h$. At each time step $h$, for any pair of $(x_h,a_h)\in \X\times\A$, which is also feasible under $M$, define $\wb Q_\theta(a_h,x_h)$ and $\wb V_\theta(x_h)$
as the standard $Q$ and value function following the policy induced by $\theta$, i.e.,
\begin{align*}
\wb V_{\theta}(x_{h}):= \E_{\pi_\theta}\Big[\sum_{h'=h}^{|\tau|} \gamma^{h'-h} r_{h'}|x_{h} \Big],~~~\wb Q_{\theta}(a_h,x_{h} ):=
\E_{\pi_\theta}\Big[\sum_{h'=h}^{|\tau|} \gamma^{h'-h} r_{h'}|x_{h},a_{h} \Big].
\end{align*}
To ease the reading, we drop the necessary conditioning on the event $h\leq |\tau|$ in our notation, while keeping its importance in mind. For a given policy $\pi\in\calP$ induced by $\theta$, define $\mu_\theta$ as its representative policy or action\footnote{In this paper, we assume that representative policies live in a Borel spaces on finite dimensional Euclidean spaces.}, also known as randomized decision rule~\citep{puterman2014markov}, on the hidden states, i.e., for any $(x_h,a_h)$ pair and all $h$, 
\begin{align*}
    \mu_\theta(a_h|x_h ) := \int_{y_h} O(y_h|x_h)\pi_\theta(a_h|y_h)   dy_h.
\end{align*}
Note that applying the above bounded linear operator on $\pi_\theta$ results in $\mu_\theta(x_h)$ which is the action probability distribution at state $x_h$. For any time step $h$, and  $x_h\in X$, consider the space of all action probability distribution generated by $\mu_\theta(x_h )$ when sweeping $\theta$ over $\Theta$, i.e., range of the mentioned linear operator when applied on $\calP$. Let $\wb a_h$ denote a point in this space. For any time step $h$, we define $\wt Q_{\theta}$, the $Q$ value of committing $\wb a_h$ at state $x_h$, and following $\pi_\theta$ afterward. More formally, at a time step $h$, state $x_h$,  and $\theta'\in\Theta$, consider $\wb a_h = \mu_{\theta'}(x_h)$, 
\begin{align*}
\wt Q_\theta(\wb a_h,x_h):=&\int_{a_h} \wb Q_\theta(a_h,x_h ) \mu_{\theta'}(a_h|x_h )  da_h=\int_{y_h,a_h} Q_\theta(a_h,x_h ) O(y_h|x_h )\pi_{\theta'}(a_h|y_h )   dy_hda_h.
\end{align*}
If we choose $\wb a_h = \mu_{\theta}(x_h)$, then $\wb V_\theta(x_h):=\wt Q_\theta(\wb a_h,x_h)|_{\wb a_h= \mu_{\theta}(x_h)}$.
Similarly, we also define the corresponding expected reward and the transition kernel under $\wb a_h = \mu_{\theta'}(x_h)$ as follows,\footnote{For an \MDP based such definitions, please refer to \citet{puterman2014markov}, e.g., Eq.~ 2.3.2.}

\begin{align*}
    \wb r(\wb a_h,x_h)&:=\int_{y_h,a_h} \wb R(a_h,y_h ) \pi_{\theta'}(a_h|y_h )O(y_h|x_h )  dy_hda_h,\\
    \wb T(x_{h+1}|\wb a_h,x_h)&:=\int_{y_h,a_h}  T(x_{h+1}|a_h,x_h ) \mu_{\theta'}(a_h|x_h )  da_h,
\end{align*}

Therefore, for any $\wb a_h$, the dynamic program on the underlying latent states is,
\begin{align*}
    \wt Q_\theta(\wb a_h, x_h ) =  \wb r(\wb a_h,x_h )+\gamma \int_{x_{h+1}} \wb T(x_{h+1}|\wb a_h,x_h ) \wb V_\theta(x_{h+1})dx_{h+1}.
\end{align*}

Now we extend the main PG theorems in \citet{sutton2000policy} and \citet{silver2014deterministic} to episodic \POMDP{}s, derived by simple modifications in their proof predecessors. 
\begin{theorem}[Policy Gradient]\label{thm:PolicyGradeint}
For a given policy $\pi_\theta$ on a \POMDP $M$;
\begin{align*}
    \Gr \eta(\theta) &= \int_{\tau\in\Upsilon}\sum_{h=1}^{|\tau|}\gamma^{h-1}f(\tau_{1..h-1},x_h,y_h;\theta)\Gr\pi_\theta(a_h|y_h) \wb Q_\theta(a_h,x_h)d\tau,
\end{align*}
and:
\begin{align*}
\Gr \eta(\theta) &=
\int_{\tau\in\Upsilon}\sum_{h=1}^{|\tau|}\gamma^{h-1} f(\tau_{1..h};\theta)\Gr\log\left(\pi_\theta(a_h|y_h)\right) \wb Q_\theta(a_h,x_h)d\tau.
\end{align*}
Proof of the Thm~\ref{thm:PolicyGradeint} in Subsection~\ref{proof:PolicyGradeint}.
\end{theorem}

\begin{theorem}[Deterministic Policy Gradient]\label{thm:DeterministicPolicyGradeint}
For a given policy $\mu_\theta$ on a \POMDP $M$;
\begin{align*}
    \Gr \eta(\theta) =\int_{\tau\in\Upsilon}\sum_{h=1}^{|\tau|}\gamma^{h-1}f(\tau_{1..h-1},x_h;\theta)\Gr \wb\mu_\theta(x_h)^\top \nabla_{\wb a_h} \wt Q_\theta(\wb a_h,x_h)|_{\wb a_h =\wb\mu_\theta(x_h)}d\tau 
\end{align*}
Proof of the Thm~\ref{thm:DeterministicPolicyGradeint} in Subsection~\ref{proof:DeterministicPolicyGradeint}.

\end{theorem}

\begin{remark} By setting the emission kernel $O$ to the identity map, the Thms.~\ref{thm:PolicyGradeint} and \ref{thm:DeterministicPolicyGradeint} reduce to their \MDP cases.
\end{remark}
These results show how the knowledge of the underlying state can help to compute the gradient of $\Gr \eta(\theta)$. Of course in practice, under partial observability circumstances, we do not have access to the latent state, unless we consider the whole history as our current observation. We provided the above mentioned results, for the sake of completeness, and latter we utilize their statements, proofs, and derivations.

\subsection{Advantage-based  Policy Gradient Theorem}
In the following, we denote $\pi_{\theta}$ as the \textit{current policy} which is the policy that we run to collect trajectories and $\pi_{\theta'}$ as a \textit{new policy} that we are interested in evaluating its performance. 
Generally, when a class of Markovian policies is considered for \POMDP{}s, neither $Q$ nor $V$ functions on observation-action pairs are well-defined as they are for \MDP{}s through the Bellman equation. In the following, we define two new quantities, similar to the $Q$ and $V$ of \MDP{}s, for \POMDP{}s. We adopt the same notation as $Q$ and $V$ for these two new quantities.

\begin{definition}[$V$ and $Q$ values in \POMDP{}s ]
For a given \POMDP $M$, the $V$ and $Q$ functions of a Markovian policy $\pi$ are as follows:
\begin{align}\label{eq:Q-V}
V_\pi(y_{h} ,a_{h-1},y_{h-1}):=&
\E_\pi\Big[\sum_{h'=h}^{|\tau|} \gamma^{h'-h} r_{h'}|y_h,y_{h-1},a_{h-1}\Big],\nonumber\\
Q_\pi(y_{h+1},a_{h},y_{h} ):=&
\E_\pi\Big[\sum_{h'=h}^{|\tau|} \gamma^{h'-h} r_{h'}| y_{h+1},y_h, a_{h}\Big].
\end{align}
For $h=0$ we relax the notation on $y_{h-1}$ in $V_\pi$ and simply denote it as $V_\pi(y_1,-,-)$ where the symbol $-$ is a member of neither the set $\Y$ nor the set $\A$.
\end{definition}


Now consider the special case where the observation is equal to the latent state, and reduce the problem to an \MDP. 
In this case we denote the value and $Q$ functions as $V_\pi(x_{h} )=\wb V_\pi(x_{h} )$, and $Q_\pi(a_h,x_{h} ) =\wb Q_\pi(a_h,x_{h} ) $.
%
For a given \MDP, the advantage function~\citep{baird1993advantage} of a policy $\pi$ is defined as the following,  for all $h$, $x_h$, and $a_h$,
\begin{align*}
A_{\pi}(x_h,a_h):=  Q_\pi(a_h,x_{h} ) -V_\pi(x_{h} ).
\end{align*}
Similarly, such advantage function is not well-defined on observations in \POMDP{}s. In the following, we define a new quantity for \POMDP{}s and call it again advantage function. We also adopt the $A$ notation for the advantage function of \POMDP{}s.

\begin{definition}[Advantage function in \POMDP{}s ]
For a given \POMDP, the advantage function of a Markovian policy $\pi$ on any tuple $\lbrace y_{h+1},a_{h},y_h,a_{h-1} ,y_{h-1}\rbrace$  is defined as follows,
\begin{align*}
&A_\pi(y_{h+1},a_{h},y_h,a_{h-1} ,y_{h-1})=Q_\pi(y_{h+1},a_h,y_{h} )-V_\pi(y_{h},a_{h-1},y_{h-1}).
\end{align*}
Similar to the definition of the value function $V$, for $h=0$ we relax the notation on $y_{h-1}$ in $A_\pi$ and simply denote it as $A_\pi(y_{2},a_{1},y_1,-,-)$
\end{definition}
These choice of such value, Q, and A function becomes clear in the following.


\begin{lemma}\label{lem:improvment}
The difference between the expected return of a policy $\pi_{\theta'}$ and of a policy $\pi_{\theta}$, denoted as improvement in the expected return is as follows,
\begin{align*}
\eta(\pi_{\theta'})-\eta(\pi_{\theta}) = \E_{\tau\sim\pi_{\theta'}}&\sum_{h=1}^{|\tau|}\gamma^{h-1} A_{\pi_{\theta}}(y_{h+1},a_h,y_h ,a_{h-1},y_{h-1}).
\end{align*}
Proof of the \lem~\ref{lem:improvment} in Subsection~\ref{proof:improvment}.
\end{lemma}

\begin{remark}
The \lem\ref{lem:improvment} is the \POMDP generalization of \lem 6.1. in \citet{kakade2002approximately} which is for \MDP{}s. This \lem also recovers the \lem 6.1. in \citet{kakade2002approximately} if the environment is in fact an \MDP.
\end{remark}

For any time step $h$, state $x_h$,  and $\theta',\theta\in\Theta$, consider $\wb a_h = \mu_{\theta'}(x_h)$, we further define, 
\begin{align*}
    \wt A_{\pi(\theta)}(x_h,\wb a_h)|_{\wb a_h=\mu_{\theta'}(x_h)}:= \wb V_\pi(x_{h} )- \wt Q_\pi(\wb a_h,x_{h} )|_{\wb a_h=\mu_{\theta'}(x_h)}
\end{align*}
Having $\wt A$ defined, we have,
\begin{corollary}\label{cor:improvment}
Given an \POMDP $M$, two policy parameters $\theta',\theta\in\Theta$,  and action representation policy of $\theta'$, i.e., $\wb a_h = \mu_{\theta'}(x_h)$, we have
\begin{align*}
\eta(\pi_{\theta'})-\eta(\pi_{\theta}) = \E_{\tau\sim\pi_{\theta'}}&\sum_{h=1}^{|\tau|}\gamma^{h-1} \wt A_{\pi(\theta)}(x_h,\wb a_h)|_{\wb a_h=\mu_{\theta'}(x_h)}
\end{align*}
\end{corollary}

The equality derived in \lem\ref{lem:improvment} suggests that if we have the advantage function of the current policy $\pi_{\theta}$ and sampled trajectories from $\pi_{\theta'}$, we could compute the improvement in the expected return $\eta(\pi_{\theta'})-\eta(\pi_{\theta})$. Then we could maximize this quantity over $\theta'$, or potentially, directly maximize the expected return $\eta(\pi_{\theta'})$ without incorporating $\pi_{\theta}$, to find an optimal policy. But in practice, we mainly do not have sampled trajectories for all new policies $\pi_{\theta'}$ to accomplish the maximization task. Instead, we have sampled trajectories from the current policy, $\pi_{\theta}$, and bound to the best use of them. To address this gap, we defined the following surrogate objective function,
\begin{align}\label{eq:surrogate}
&L_{\pi_{\theta}}(\pi_{\theta'}):=\eta(\pi_{\theta})+\E_{\tau\sim\pi_{\theta}}\sum_{h=1}^{|\tau|}\gamma^{h-1}\E_{a'_h\sim \pi_{\theta'}(\cdot|y_h)}A_{\pi_{\theta}}(y_{h+1},a'_h,y_h ,y_{h-1},a_{h-1}).
\end{align}
One can compute this surrogate objective using sampled trajectory generated by following the current policy $\pi_\theta$, advantage function induced by the same policy $\pi_\theta$, and evaluating the advantage function by choosing actions $a'_h$ according to $\pi_{\theta'}$. Therefore, we can maximize the surrogate objective function Eq.~\ref{eq:surrogate} over $\theta'$ just using sampled trajectories and advantage function of the current policy $\pi_{\theta}$. The question is how maximizing or increasing $L_{\pi_{\theta}}(\pi_{\theta'})$ is related to our quantity of interest $\eta(\pi_{\theta'})$. In the following, we study this relationship. The gradient of $L_{\pi_{\theta}}(\pi_{\theta'})$ with respect to $\theta'$, then evaluated at $\theta$ is expressed as following,
\begin{align}\label{eq:PolicyGradeintSurrogate}
    \nabla_{\theta'}L_{\pi_{\theta}}(\pi_{\theta'})|_{\theta'=\theta}&=\nabla_{\theta'}\Big(\eta(\pi_{\theta})+\E_{\tau\sim\pi_{\theta}}\sum_{h=1}^{|\tau|}\gamma^{h-1}\E_{a_h'\sim \pi_{\theta'}(a_h|y_h)}A_{\pi_{\theta}}(y_{h+1},a_h',y_h ,a_{h-1},y_{h-1})\Big)|_{\theta'=\theta}\nonumber\\
    &\!\!\!\!\!\!\!\!\!\!\!\!\!\!\!\!\!\!\!\!\!\!\!\!=\int_{\tau\in\Upsilon}\sum_{h=1}^{|\tau|} \gamma^{h-1} f(\tau_{1..h},x_{h+1},y_{h+1};\theta)\Big(\int_{a'_h}\nabla_{\theta'} \pi_{\theta'}(a'_h|y_h)|_{\theta'=\theta} A_{\pi_{\theta}}(y_{h+1},a_h',y_h,a_{h-1},y_{h-1})da'_h\Big)d\tau.
\end{align}
We later use Eq.~\ref{eq:PolicyGradeintSurrogate} for the convergence analysis of \PG methods.
For any pairs of two policies, $\pi_{\theta}$ and $\pi_{\theta'}$, and any time step $h$ and observations $y_h$, consider $\check a_h = \pi_{\theta'}(y_h)$ as the representative action of $\pi_{\theta'}$ at observation $y_h$,  we have

\begin{align}\label{eq:tildA}
A_{\pi_{\theta}}(y_{h+1},\check a_h,y_h , a_{h-1},y_{h-1})
&=\int_{a_h} A_{\pi_{\theta}}(y_{h+1},a_h,y_h ,a_{h-1},y_{h-1})\pi_{\theta'}(a_h|y_h) da_h.
\end{align}

Note that we are using the same notation $A$ for both primitive actions $a_h$ and representative actions $\check a_h$.
%
%
%

\begin{theorem}[Surrogate Deterministic Policy Gradient]\label{thm:DeterministicPolicyGradeintMemory}
Given a \POMDP $M$, and the advantage function induced by policy $\pi_\theta$, the gradient of the surrogate objective of $\pi_{\theta'}$, $\nabla_{\theta'}L_{\pi_{\theta}}(\pi_{\theta'})$, with respect to $\theta'$, evaluated at $\theta$ is as follows:
\begin{align*}
\nabla_{\theta'}L_{\pi_{\theta}}(\pi_{\theta'})|_{\theta'=\theta}&
\\
&\!\!\!\!\!\!\!\!\!\!\!\!\!\!\!\!\!\!\!\!=\int_{\tau\in\Upsilon}\sum_{h=1}^{|\tau|} \gamma^{h-1} f(\tau_{1..h},x_{h+1},y_{h+1};\theta)\nabla_{\theta'} \pi_{\theta'}(y_h)|_{\theta'=\theta}^\top \nabla_{\check a_h} A_{\pi_{\theta}}(y_{h+1},\check a_h,y_h,a_{h-1},y_{h-1})|_{\check a_h = \pi_{\theta}(y_h)}d\tau
\end{align*}
Proof of the \thm\ref{thm:DeterministicPolicyGradeintMemory} in Subsection~\ref{proof:DeterministicPolicyGradeintMemory}.
\end{theorem}
Note that when the problem is fully observable and \MDP{}s, marginalizing over $x_t+1$ and $y_t+1$ at each time step $h$ in the theorem statement reduces the gradient of surrogate function to its \MDP{} case. In \MDP setting, we have $\nabla_{\theta'}L_{\pi_{\theta}}(\pi_{\theta'})|_{\theta'=\theta}=\nabla_{\theta'}\eta_(\pi_{\theta'})|_{\theta'=\theta}$, which is one the key component in analyses of \PG methods in \MDP{}s~\citep{kakade2002approximately,schulman2015trust}. 

\subsection{Gradient Dominance and Surrogate Function}
In the following we study the convergence properties of \PG methods on surrogate function $L_{\pi_{\theta}}(\pi_{\theta'})$. Consider the following operator, as an alternative to Bellman max operator: 
\begin{definition}\label{Def:opt_operator}
Given a \POMDP $M$ and an advantage function $A_{\pi_{\theta}}$, define the optimizing operator $\T$ as follows:
\begin{align*}
    \T A_{\pi_{\theta}}= &\max_{\theta' }\int_{\tau}\sum_{h=1}^{|\tau|}\gamma^{h-1} f(\tau_{1..h},x_{h+1},y_{h+1};\theta)\\
    &\left(A_{\pi_{\theta}}(y_{h+1},\pi_{\theta'}(y_h),a_h,y_h ,y_{h-1},a_{h-1})-A_{\pi_{\theta}}(y_{h+1},\pi_{\theta}(y_h),a_h,y_h ,y_{h-1},a_{h-1})\right)d\tau.
\end{align*}
with $\theta^\star(\theta)$ a maximizer and, $\pi_{\theta^\star(\theta)}$, its corresponding policy.
\end{definition}
$\T A_{\pi_{\theta}}$ represents maximum improvement one can make on $L_{\pi_{\theta}}(\pi_{\theta'})$ by changing $\theta'$. Consider the following set of assumptions.
\begin{assumption}\label{asm:convex}
For every time step $h$, and tuples of $y_{h+1},y_h ,y_{h-1},a_{h-1}$, $A_{\pi_{\theta}}(y_{h+1},\check a_h,y_h ,y_{h-1},a_{h-1})$ is a concave function of $\check a_h$. 
\end{assumption}


\begin{assumption}\label{asm:existence}
For any $\theta$, there exist a vector $\varpi(\theta)$ in the space of observation representative policies, such that for any time step $h$ and observation $y_h$,   $\frac{\partial}{\partial\alpha}\pi_{\theta+\alpha \varpi(\theta)}(y_h)|_{\alpha = 0} = \pi_{\theta^\star(\theta)}(y_h)-\pi_\theta(y_h)$.
\end{assumption}

\begin{theorem}[Convergence of \PG on Surrogate Objective]\label{thm:Convergence}
Given a \POMDP $M$, under the Asms.~\ref{asm:convex}, and~\ref{asm:existence}, for a policy $\pi_\theta$, and a corresponding function $\varpi(\theta)$, we have,
\begin{align*}
    \T A_{\pi_{\theta}}\leq \nabla_{\alpha}L_{\pi_{\theta}}(\pi_{\theta+\alpha \varpi(\theta)})|_{\alpha=0}.
\end{align*}

Proof of the \thm\ref{thm:Convergence} in Subsection~\ref{proof:Convergence}.
\end{theorem}

\begin{corollary}[Optimally and Surrogate Objective]
Since $\T A_{\pi_{\theta}}$ is always non negative, when the gradient $\nabla_{\theta'}L_{\pi_{\theta}}(\pi_{\theta'})|_{\theta'=\theta}$, is zero, the $\T A_{\pi_{\theta}}$ is equal to zero.
\end{corollary}
The \thm\ref{thm:Convergence} is the \POMDP generalization of the main theorem in \citet{bhandari2019global}, which is developed for \MDP{}s. In the case of full observable setting, the \thm\ref{thm:Convergence} reduces to its \MDP counterpart and can be seen by marginalizing $x_{h+1}$ and $y_{h+1}$ for each times step $h$ in the derivation. The deployed assumptions in the statement of the \thm\ref{thm:Convergence} are limiting, but the resulting statement is strong. In the following we relax these assumptions.

\subsection{Convergence Guarantee and Distribution Mismatch Coefficient}
In the following we study the converges guarantee of \PG methods without the assumptions in the previous subsection. We show that convergence takes place in presence of coefficients related to the richness of policy class, and how far the underlying \POMDP is from being an \MDP.
\begin{definition}[Distribution mismatch coefficient]
Given any pairs of parameters $\theta$ and $\theta'$, the infinity norm of ratio of measures $f(\tau:\theta)d\tau$, and $f(\tau:\theta')d\tau$,\footnote{In general, it need not to be with respect to Lebesgue.} denotes the distribution mismatch coefficient between $\pi_\theta$ and $\pi_{\theta'}$, i.e.,
$\|\frac{f(\tau:\theta')}{f(\tau:\theta)}\|_\infty$
with respect to the measure space $(\Upsilon,\F,\PP_{\theta'})$.
\end{definition}
The distribution mismatch coefficient for $(\theta,\theta')$ express how representative the trajectories generated by following $\pi(\theta)$ are for the trajectories generated by following $\pi(\theta')$. In the following we use this notation for $(\theta,\theta^*)$, i.e., $\|\frac{f(\tau:\theta^*)}{f(\tau:\theta)}\|_\infty$, on how representative are samples of $\pi(\theta)$ for $\pi(\theta^*)$. The definition of distribution mismatch coefficient up to a slight modification is an \POMDP{} generalization its \MDP counterpart~\citep{antos2008learning,agarwal2019optimality}.
%
%
For a given parameter $\theta$ and an optimal parameter $\theta^*$, let $\theta^+\in\Theta$ denote a maximizer to the following optimization,\footnote{One should use $\theta^+(\theta,\theta^*)$ notation, but we adopt $\theta^+$ for simplicity since we are defining it for a given $\theta,\theta^*$.}
\begin{align}\label{eq:policyplus0}
    \theta^+\in& \argmax_{\theta''\in\Theta}\int_{\tau}\sum_{h=1}^{|\tau|} \gamma^{h-1} f(\tau_{1..h-1},x_{h},y_{h};\theta^*)\pi_{\theta''}(a_h|y_h)\nonumber\\
    &\quad\quad\quad\quad\quad\quad\quad T(x_{h+1}|x_h,a_h)O(y_{t+1}|x_{t+1})A_{\pi_{\theta}}(y_{h+1},a_h,y_h ,a_{h-1},y_{h-1}) d\tau.
\end{align} 
Note that the objective on this optimization is positive since setting  $\pi_{\theta''}=\pi_{\theta*}$ reduces the right hand side to $\eta(\theta^*)-\eta(\theta)$, and therefore, if $\theta=\theta^*$, we have $\eta(\theta^*)=\eta(\theta^+)$. Considering the actions of policy $\pi_{\theta^+}$ on trajectories generated using $\tau\sim f(\cdot;\theta)$, we define an indicator function $e(x_h,a_{h-1},y_{h-1})$, such that for each $h$, and $x_h,a_{h-1},y_{h-1}$, $e(x_h,a_{h-1},y_{h-1})$ is equal to $1$ if,
\begin{align*}
    \int_{y_{h},a_h,x_{h+1},y_{h+1}}\!\!\!\!\!\!\!\!\!\!\!\!\!\!\!\!\!\!\!\!\!\!\!\!\!\!\!\!O(y_{t}|x_{t})\pi_{\theta^+}(a_h|y_h)  T(x_{h+1}|x_h,a_h)O(y_{h+1}|x_{h+1})A_{\pi_{\theta}}(y_{h+1},a_h,y_h ,a_{h-1},y_{h-1}) dy_{h}da_hdx_{h+1}dy_{h+1} \geq 0,
\end{align*}
and zero otherwise. In tabular \MDP{}s, with tabular policies, since one can choose actions that make the evaluated advantage function at each state to be positive, $e(x_h,a_{h-1},y_{h-1})=e(x_h,a_{h-1},x_{h-1})$ is always equal to one, which follows by the definition of $\theta^+$. For each tuple of $(a_h,y_h,x_h,a_{h-1},y_{h-1})$, we define $\pi^+_{\theta^+,\theta}$ which keep parts of trajectories that contributed in positive increments in Eq.~\ref{eq:policyplus0}:
\begin{align}\label{eq:piplus}
\pi^+_{\theta^+,\theta}(a_h,y_h,x_h,a_{h-1},y_{h-1}):= 
 e(x_h,a_{h-1},y_{h-1})\pi_{\theta^+}(a_h|y_h),
\end{align}
where $\pi^+_{\theta^+,\theta}$, while being a kernel from the measure theoretic point of view, it might not necessarily be a probability kernel. We note that $\pi^+_{\theta^+,\theta}$ reduced to $\pi(\theta^+)$ in tabular \MDP{}s. Therefore the difference between $\pi^+_{\theta^+,\theta}$ and $\pi(\theta^+)$ accounts for both partial observably of the system, as well as the expressively of the function class used for polices. In the following we consider the performance of $\pi(\theta)$ and how much $\pi^+_{\theta^+,\theta}$ improves this performance: 
\begin{align}\label{eq:supAdvant30}
\Delta\eta^+(\theta^+,\theta):=&\int_{\tau}\sum_{h=1}^{|\tau|} \gamma^{h-1} f(\tau_{1..h-1},x_{h},y_{h};\theta)\pi^+_{\theta^+,\theta}(a_h,y_h,x_h,a_{h-1},y_{h-1})\nonumber\\
&\quad\quad\quad\quad\quad\quad\quad\quad\quad T(x_{h+1}|x_h,a_h)O(y_{h+1}|x_{h+1})A_{\pi_{\theta}}(y_{h+1},a_h,y_h ,a_{h-1},y_{h-1}) d\tau\nonumber\\
&-\int_{\tau}\sum_{h=1}^{|\tau|} \gamma^{h-1} f(\tau_{1..h},x_{h+1},y_{h+1};\theta)A_{\pi_{\theta}}(y_{h+1},a_h,y_h ,a_{h-1},y_{h-1}) d\tau\nonumber\\
&=\int_{\tau}\sum_{h=1}^{|\tau|} \gamma^{h-1} f(\tau_{1..h-1},x_{h},y_{h};\theta)\left(\pi^+_{\theta^+,\theta}(a_h,y_h,x_h,a_{h-1},y_{h-1})-\pi_{\theta}(a_h|y_h) \right)\nonumber\\
&\quad\quad\quad\quad\quad\quad\quad\quad T(x_{h+1}|x_h,a_h)O(y_{h+1}|x_{h+1})A_{\pi_{\theta}}(y_{h+1},a_h,y_h ,a_{h-1},y_{h-1}) d\tau.
\end{align}
Using these pieces, finally, we define Bellman policy error $\epsilon^+_{\theta}$, and its corresponding compatibility vector, ${\omega^+}$, two problem dependent quantities in Def.~\ref{def:BPE}. These parameters summarize  the expressivity of the policy class, and the closeness of the \POMDP to an \MDP behavior.

\begin{definition}[Bellman policy error]\label{def:BPE}
Bellman policy error $\epsilon^+_{\theta}$ is the error of a best linear approximation under mean absolute error to $\Delta\eta^+(\theta^+,\theta)$ under feature vector $\nabla_\theta\pi_{\theta}(a_h|y_h)\in \Re^d$ of observation-action pairs,
\begin{align*}
    &\epsilon^+_{\theta}:= \min_{\omega\in\Re^d}\Big|\Delta\eta^+(\theta^+,\theta)-
    {\omega^+}^\top\E_{\tau\sim\pi(\theta)}\sum_{h=1}^{|\tau|} \gamma^{h-1} \big(\!\!\int_{a'_h}\!\!\nabla_\theta\pi_{\theta}(a_h'|y_h) A_{\pi_{\theta}}(y_{h+1},a_h',y_h ,a_{h-1},y_{h-1})da'_h\big)\Big|,
\end{align*}
where $\omega^+_{\theta}$, a compatibility vector of Bellman policy error, as a minimizer. 
\end{definition}{}
Since there might be multiple $\pi_{\theta^+}$ as a result of Eq.~\ref{eq:policyplus0} maximization, one can consider a $\pi_{\theta^+}$ which minimizes $\epsilon^+_{\theta}$. Using the approximation error $\epsilon^+_{\theta}$, and the distribution mismatch coefficient, we have the following theorem.

\begin{theorem}[Gradient dominance and distribution mismatch coefficient]\label{thm:Convergence_dist}
Given a \POMDP $M$, parameter set $\Theta$ and the corresponding measure space $(\Gamma,\Upsilon,\PP_\theta)$, for any $\theta$ we have,
\begin{align*}
    &\eta(\theta^*)-\eta(\theta) \leq\epsilon^+_{\theta}\|\frac{f(\tau:\theta^*)}{f(\tau:\theta)}\|_\infty+\|\frac{f(\tau:\theta^*)}{f(\tau:\theta)}\|_\infty{\omega^+}^\top\nabla_{\theta'}L_{\pi_{\theta}}(\pi_{\theta'})|_{\theta'=\theta}.
\end{align*}
\end{theorem}
Proof of the \thm\ref{thm:Convergence_dist} in Subsection~\ref{proof:Convergence_dist}.

\begin{corollary}[Optimally and Bellman Policy Error]
The \thm\ref{thm:Convergence_dist} indicates that following policy gradient on the surrogate objective $L_{\pi_{\theta}}(\pi_{\theta'})$ to the point that either the gradient $\nabla_{\theta'}L_{\pi_{\theta}}(\pi_{\theta'})|_{\theta'=\theta}$ is a vector with a small norm, policies are compatible (i.e., small norm on ${\omega^+}$), or small value on their inner product, and importantly small approximation error $\epsilon^+(\theta)$, as long as the distribution mismatch is finite, then the performance of $\pi(\theta)$ is close to that of $\pi(\theta^*)$.
\end{corollary}
The results in the \thm\ref{thm:Convergence}, as well as the definitions of Bellman policy error and compatibility vector are the extension of the mains results, in the section 6 of the recent paper by \citet{agarwal2019optimality}. For the discussion of a similar theorem in \MDP case, please refer to the mentioned paper.


\section{Trust Region Policy Optimization}


We now study how to formally extend trust region-based policy optimization methods to \POMDP{}s.
Generally, the notion of gradient depends on the parameter metric space. 
%
Given a pre-specified Riemannian metric, a gradient direction is defined. When the metric is Euclidean, the notion of gradient reduces to the standard gradient \citep{lee2006riemannian}. 
The Riemannian generalization of the notion of gradient adjusts the standard gradient direction based on the local curvature induced by the Riemannian manifold of interest. A valuable knowledge of the curvature assists in finding an ascent direction, which might conclude to big ascend in the objective function. This approach is also interpreted as a trust region method where we are interested in assuring that the ascent steps do not change the objective beyond a safe region where the local curvature might not stay valid. In general, a valuable manifold might not be given, and we need to adopt one. Fortunately, when the objective function is an expectation over a parameterized distribution, \citet{amari2016information} recommends employing the Riemannian metric, induced by the local Fisher information. This choice of metric results in a well knows notion of the gradient, known as \textit{natural gradient}. For the objective function in the Eq.~\ref{eq:eta}, the Fisher information matrix is defined as the following,
\begin{align}\label{eq:fisher}
F(\theta):=\int_{\tau\in\Upsilon}f(\tau;\theta) \left[\nabla_\theta \log\left(f(\tau;\theta)\right)\nabla_\theta \log\left(f(\tau;\theta)\right)^\top\right] d\tau.
\end{align}
Deployment of natural gradients in \PG methods, at least goes back to \citet{kakade2002natural} in \MDP{}s.
The direction of the gradient with respect to $F$-induced metric is derived as $F(\theta)^{-1}\nabla_{\theta}(\eta(\theta))$. One can compute the inverse of this matrix to come up with the direction of the natural gradient. Since neither storing the Fisher matrix is always bearable, nor computing the inverse is practically desirable, direct utilization of $F(\theta)^{-1}\nabla_{\theta}(\eta(\theta))$ might not be a feasible option.

A classical approach to estimate $F^{-1}\nabla_{\theta}(\eta)$ is based on compatible function approximation methods. \citet{kakade2002natural} study this approach in the context of \MDP{}s. In the following, we develop this approach for \POMDP{}s. Consider a feature map $\phi(\tau)$ in a desired ambient space. We approximate the return $R(\tau)$ via a linear function of the feature representation $\phi(\tau)$, i.e.,
\begin{align*}
\min_\omega\epsilon(\omega):=\int_{\tau}f(\tau,\theta)\big(\phi(\tau)^{\top}\omega-R(\tau)\big)^2d\tau.
\end{align*}
which is a convex program with a minimizer $\omega^*$. To find the optimal $\omega$, we take the gradient of $\epsilon(\omega)$ with respect to $\omega$, and solve it for zero vector, i.e.,
\begin{align*}
\bold 0=\nabla_{\omega}\epsilon(\omega)|_{\omega=\omega^*}=
\int_{\tau}2f(\tau,\theta)\phi(\tau)[\phi(\tau)^{\top}\omega^*-R(\tau)]d\tau.
\end{align*}
For the optimality, 
\begin{align*}
\int_{\tau}f(\tau,\theta)\phi(\tau)\phi(\tau)^\top\omega^* d\tau=
\int_{\tau}f(\tau,\theta)\phi(\tau;\theta)R(\tau)d\tau.
\end{align*}
If we consider the $\phi(\tau)= \nabla_{\theta}\log \left(\Pi_{h=1}^{|\tau|}\pi_\theta(a_h|y_h)\right)
$, the LHS of this equality is $F(\theta)\omega^*$. Therefore
\begin{align*}
F(\theta)\omega=\nabla_{\theta}\eta(\theta)\Longrightarrow \omega^*=F(\theta)^{-1}\Grrho.
\end{align*}
This is one way to compute $F(\theta)^{-1}\Grrho$. 
Another approach to estimating the natural gradient relies on the computation of the KL divergence. This approach, that also has been utilized in \TRPO, suggests to first deploy  $\KL$ divergence substitution technique and then use conjugate gradient procedure to tackle the computation and storage bottlenecks of direct computation of natural gradient. In the following, we analyze this approach in the episodic settings.

\begin{lemma}\label{lem:KL-Fisher}
Under a set of mild regularity conditions,
\begin{align}\label{eq:kl}
\nabla^2_{\theta'}\KL(\theta,\theta')|_{\theta'=\theta}=F(\theta),
\end{align}
with $\KL(\theta,\theta'):=-\int_{\tau\in\Upsilon} \!f(\tau;\theta) \log\left(f(\tau;\theta')/f(\tau;\theta)\right) d\tau$, and $\nabla^2_{\theta'}$ the Hessian with respect to ${\theta'}$.
\end{lemma}
The \lem\ref{lem:KL-Fisher} is a known lemma in the literature, and we provide its proof in the subsection~\ref{proof:KL-Fisher}. In practice, it might not be feasible to compute the expectation in either the Fisher information matrix or the $\KL$, but rather their empirical estimates. Given $m$ trajectories:
\begin{align*}
\nabla^2_{\theta'} {\wKL}(\theta,\theta')|_{\theta'=\theta}=-\frac{1}{m}\nabla^2_{\theta'}\sum_{t=1}^{m}\sum_{h=1}^{|\tau^t|}\log\left(\frac{\pi_{\theta'}(a^t_h|y^t_h)}{\pi_\theta(a^t_h|y^t_h)}\right).
\end{align*}
This empirical estimate is the same for both \MDP{}s and \POMDP{}s. The analysis in most of the celebrated \PG methods, e.g., \TRPO, \PPO, are dedicated to infinite horizon \MDP{}s, while almost all the experimental studies in this line of work are in the episodic settings. Therefore the estimator used in these methods,
\begin{align*}
&\nabla^2_{\theta'}{\wKL}^{TRPO}(\theta,\theta')|_{\theta'=\theta}=-\frac{1}{\sum_t^m|\tau^t|}\nabla^2_{\theta'}\sum_{t=1}^{m}\sum_{h=1}^{|\tau^t|}\log\left(\frac{\pi_{\theta'}(a^t_h|y^t_h)}{\pi_\theta(a^t_h|y^t_h)}\right),
\end{align*}
is a biased estimator to the $\KL$ in episodic settings, which is the $\KL$ required in the mentioned studies. This bias, for example, can result in a dramatic failure of \TRPO construction of the trust region and prevent it from fulfilling its monotonic improvement promise. While this bias might results in degradation of empirical performance, of course, there are scenarios that this bias, in fact, helps to improve the performance. In this paper, we are concerned with properties that are aligned with theoretical guarantees.

One can skip the following subsection, with no serious harm in the subjects afterward. In the following subsection, we provide a discussion on $\nabla^2_{\theta'}{\KL}$ vs. $\nabla^2_{\theta'}{\KL}^{TRPO}$.



\subsection{$\KL$ vs. ${\KL}^{TRPO}$}\label{remark:kl}
The use of $ {\KL}$ instead of ${\KL}^{TRPO}$ is motivated by theory and also intuitively recommended. A small change in the policy at the beginning of short episodes does not make a drastic shift in the distribution of the trajectory but might cause radical shifts when the trajectory length is long. Therefore, for longer horizons, the trust region needs to shrink. 

For the sake of simplicity, recognize the construction of trust region as ${\KL}\leq \delta$ for some desired $\delta$. Consider two trajectories, one long and one short. The ${\KL}\leq \delta$ induces a region which allows greater changes in the policy for short trajectory while limiting changes in long trajectory. While  ${\KL}^{TRPO}\leq \delta$ induces the region, which is indifference to the length of trajectories and looks at each sample as it is experienced in a stationary distribution of an infinite horizon \MDP{}. In other words, it allows the same amount of change in policy for short episodes as it gives to long ones.

Consider a toy \RL problem where at the beginning of the learning, when the policy is not good, the agent dies at the early stages of the episodes (termination). In this case, the trust region under $ {\KL}$ is vast and allows for substantial change in the policy parameters, while again, ${\KL}^{TRPO}$ does not consider the length of the episode. On the other hand, toward the end of the learning, when the agent has learned a good policy, the length of the horizon grows, and small changes in the policy might cause drastic changes in the trajectory distribution. Therefore the trust region, under $\KL$, shrinks again, and just small changes in the policy parameters are allowed, which is again captured by ${\KL}$ but not at all by ${\KL}^{TRPO}$.

It is worth restate that one can cook up an example that ${\KL}^{TRPO}$ construction is helpful, but it is not the point of this section since it is not what the \TRPO guarantees and analysis promise. Generally, there more issues with treating episodic \RL problems as infinite horizon problems, and we refer reader to \citet{thomas2014bias} for more in this topic.

\subsection{Construction of the Trust Region} In practice, depending on the problem at hand, either of the discussed approaches for computing the natural gradient can be applicable. For the construction of trust region, one can exploit the close relationship between $\KL$ and Fisher information matrix \lem\ref{lem:KL-Fisher} and also the fact that the Fisher matrix is equal to second order Taylor expansion of $\KL$. Instead of considering the area $\|\left(\theta-\theta'\right)^\top F\left(\theta-\theta'\right)\|_2\leq \delta$, or $\|\left(\theta-\theta'\right)^\top \nabla^2_{\theta'}\KL(\theta,\theta')|_{\theta'=\theta}\left(\theta-\theta'\right)\|_2\leq \delta$ for the construction of the trust region, we can approximately consider $\KL(\theta,\theta')\leq \delta/2$. These relationships between these three approaches in constructing the trust region is used throughout this paper.

To complete the study of KL divergences, we propose a discount-factor-dependent divergence and provide the monotonic improvement guarantee with respect to $\KL$ as well as this new divergence. 

The $\KL$ divergence and Fisher information matrix in Eq.~\ref{eq:kl}, Eq.~\ref{eq:fisher} do not convey the effect of the discount factor. Consider a setting with a small discount factor $\gamma$. In this setting, we do not mind drastic distribution changes in the latter part of episodes since there achieved reward is strongly discounted anyway. Therefore, we desire to have an even wider trust region and allow bigger changes for latter parts of the trajectories. This is a valid intuition, and in the following, we derive a $\KL$ divergence by also incorporating $\gamma$. We rewrite $\eta(\theta)$ as follows,
\begin{align*}
\eta(\theta)&=\int_{\tau\in\Upsilon}f(\tau;\theta) R(\tau)d\tau= \int_{\tau\in\Upsilon} \sum_{h=1}^{|\tau|}f(\tau_{1..h};\theta) \gamma^{h-1} r_h(\tau)d\tau.
\end{align*}
Following the \citet{amari2016information} reasoning for Fisher information of each component of the sum, we derive a $\gamma$-dependent divergence $\D_\gamma(\theta,{\theta'})$,
\begin{align}\label{eq:Divergence}
  \D_\gamma(\theta,{\theta'}):=\Big(\sum_{h\geq1}\gamma^{h-1}\sqrt{\frac{1}{2}\KL\big(\tau_{1..h}\sim f(\cdot;\theta'),\tau_{1..h}\sim f(\cdot;\theta)\big)}
\Big)^2.
\end{align}
This divergence penalizes the distribution mismatch less in the latter parts of trajectories through discounting them. Similarly, taking into account the relationship between KL divergence and Fisher information we also define $\gamma$-dependent Fisher information, $F_{\gamma}(\theta)$,
\begin{align*}
F_{\gamma}(\theta):=&\int_{\tau\in\Upsilon} \sum_{h=1}\gamma^hf(\tau_{1..h};\theta)
\left[\nabla_\theta \log\left(f(\tau_{1..h};\theta)\right)\nabla_\theta \log\left(f(\tau_{1..h};\theta)\right)^\top\right] d\tau.
\end{align*}

We develop our trust region study upon both definitions of $\D$ and $\D_\gamma$.

\subsection{Generalized Trust Region Policy Optimization}
We propose Generalized Trust Region Policy Optimization (\GTRPO), a generalization of \MDP{}-based trust region methods in \citet{kakade2002approximately,schulman2015trust} to episodic environments, agnostic to whether \MDP{} or \POMDP{}. We utilize the proposed notion of advantage function, prove the monotonic improvement properties of \GTRPO, and show how the KL divergences, $\D_\gamma$ and $\KL$, play their roles in the construction of trust regions.

\begin{algorithm}[t]
\caption{\GTRPO}
\begin{algorithmic}[1]
\STATE Initial $\pi_{\theta}$, $\epsilon'$, $\delta'$
\STATE Choice of divergence $\D$: $\KL$ or $\D_\gamma$
    \FOR{epoch $=$ 1 until convergence}
        \STATE Deploy policy $\pi_\theta$ and collect experiences
        \STATE Estimate the advantage function $\wh{A}_{\pi_{\theta}}$
        \STATE Construct empirical estimates of the surrogate objective $\wh L_{\pi_{\theta}}(\pi_{\theta'})$ and $\wh\D$
        \STATE Update $\theta$ using the  
        \begin{align*}
            &\theta\in \arg\max_{\theta'}\wh L_{\pi_{\theta}}(\pi_{\theta'})~~s.t.~\frac{1}{2}\|\left(\theta'-\theta\right)\!^\top \nabla^2_{\theta''}\wh\D(\theta,\theta'')|_{\theta''=\theta}\left(\theta'-\theta\right)\|_2\leq\delta'
        \end{align*}
    \ENDFOR
\end{algorithmic}
\label{alg:trpo}
\end{algorithm}

%
%
%
%
%
%
As illustrated in Alg.~\ref{alg:trpo}, \GTRPO employs its current policy $\pi_{\theta}$ to collect trajectories, estimate the advantage function using collected samples. \GTRPO deploys the estimated advantage function to derive the surrogate objective. In order to come up with a  new policy, \GTRPO maximizes the surrogate objective over policy parameters in the vicinity of the current policy, defined through the trust region. The underlying procedures in \GTRPO are similar to its predecessor \TRPO except, instead of maximizing the surrogate objective defined over unobserved hidden state dependent advantage function, $A_{\pi_{\theta}}(a_h,x_h)$ , it maximizes the surrogate objective defined using $A_{\pi_{\theta}}(y_{h+1},a_h,y_h,y_{h-1},a_{h-1})$. 

\begin{sloppypar}
Note that if the underlying environment is an \MDP,  $A_{\pi_{\theta}}(y_{h+1},a_h,y_h,y_{h-1},a_{h-1})$ is equivalent to $A_{\pi_{\theta}}(x_{h+1},a_h,x_h)$ where after marginalizing out $x_{h+1}$ in the expectation we end up with $A_{\pi_{\theta}}(a_h,x_h)$ and recover \TRPO.
In practice, one can estimate the advantage function $A_{\pi_\theta}(y_{h+1},a,y_h,y_{h-1},a_{h-1}) $ by approximating $Q_{\pi_\theta}(y_{h+1},a,y_{h})$ and $V_{\pi_\theta}(y_{h},y_{h-1},a_{h-1})$ using data collected by following ${\pi_\theta}$ and function classes of interest. 
\end{sloppypar}

%
%
In the following we show that maximizing $L_{\pi_{\theta}}(\pi_{\theta'})$ over $\theta'$ results in a lower bound on the improvement $\eta(\pi_{\theta'})-\eta(\pi_{\theta})$ when $\pi_{\theta}$ and $\pi_{\theta'}$ are close under $\KL$ or $\D_\gamma$. Consider the maximum spans of advantage function, $\epsilon$ and $\epsilon'$, such that for all $\theta\in\Theta$, the following inequalities hold,
\begin{align*}
&\epsilon\geq\max_{\tau\in\Upsilon}\big|\sum_h^{|\tau|}\gamma^{h-1}{A}_{\pi_{\theta}}(y_{h+1},\pi_{\theta}(y_{h}),\mu_{\theta}(y_{h}),y_{h},y_{h-1},a_{h-1})\big|,\\
&\epsilon'\geq\max_{\tau\in\Upsilon}\big|A_{\pi_{\theta}}(y_{h+1},\pi_{\theta}(y_{h}),y_{h},y_{h-1},a_{h-1})\big|.
\end{align*}
Using the quantifies of $\epsilon$ and $\epsilon'$, we have the following monotonic improvement guarantees,
\begin{theorem}[Monotonic Improvement Guarantee]\label{thm:monotonic}
For two $\pi_{\theta}$ and $\pi_{\theta'}$, construct $L_{\pi_{\theta}}(\pi_{\theta'})$, then
\begin{align*}
1)\eta(\pi_{\theta'})&\geq L_{\pi_{\theta}}(\pi_{\theta'})-\epsilon TV\left(\tau\sim f(\cdot;\pi_{\theta'}),\tau\sim f(\tau;\pi_{\theta})\right)-\wb\epsilon\\
&\geq L_{\pi_{\theta}}(\pi_{\theta'})-\epsilon \sqrt{\frac{1}{2}\KL\left(\pi_{\theta'},\pi_{\theta}\right)}-\wb\epsilon,\\
2)\eta(\pi_{\theta'})&\geq L_{\pi_{\theta}}(\pi_{\theta'})-\epsilon'\sum_{h\geq1}\gamma^{h-1}\sqrt{\frac{1}{2}\KL\left(\tau_{1..h}\sim f(\cdot;\pi_{\theta'}),\tau_{1..h}\sim f(\cdot;\pi)\right)}-\wb\epsilon\\
&\geq L_{\pi_{\theta}}(\pi_{\theta'})-
\epsilon'\sqrt{\D_\gamma\left(\pi_{\theta},\pi_{\theta}\right)}-\wb\epsilon,
\end{align*}
where $\wb\epsilon$ is the advantage gap~Eq.\ref{eq:adv_gap} and is zero in case \MDP{}s.
\end{theorem}
Proof of the \thm\ref{thm:monotonic} in Subsection~\ref{proof:monotonic}.
The \thm\ref{thm:monotonic} recommends optimizing $ L_{\pi_{\theta}}(\pi_{\theta'})$ over $\pi_{\theta'}$ in the vicinity of $\pi_{\theta}$, defined by $\KL$ or $\D_{\gamma}$. More formally, \thm\ref{thm:monotonic} results suggest to maximize the shifted lower bound on $\eta(\pi_{\theta'})$, i.e., either,
\begin{align*}
L_{\pi_{\theta}}(\pi_{\theta'})-\epsilon \sqrt{\frac{1}{2}\KL\left(\pi_{\theta'},\pi_{\theta}\right)},\textit{or}~L_{\pi_{\theta}}(\pi_{\theta'})-
\epsilon'\sqrt{\D_\gamma\left(\pi_{\theta},\pi_{\theta}\right)},
\end{align*}
and accept the new parameter if either of these objective values is above the lowered expected return of the current parameter, $\eta(\theta)-\wb\epsilon$, resulting in monotonic improvement. Using the relationship between the KL divergence and Fisher information, as well as the applications of interest in practice, we propose the following alternative optimization procedures. In practice, given the current policy $\pi_{\theta}$, one might be interested in either of the following optimization,
\begin{align*}
\max_{\theta'} L_{\pi_{\theta}}(\pi_{\theta'})-C\sqrt{\KL\left(\pi_{\theta},\pi_{\theta'}\right)},~~~\textit{or}~~~
\max_{\theta'} L_{\pi_{\theta}}(\pi_{\theta'})-C'\sqrt{\D_\gamma\left(\pi_{\theta},\pi_{\theta'})\right)},
\end{align*}
where $C$ and $C'$ are the problem and application dependent constants. We also can view the $C$ and $C'$ as the knobs to restrict the trust region denoted by $\delta$, $\delta'$ and construct the following constraint optimization problems,
\begin{align*}
\max_{\theta'} L_{\pi_{\theta}}(\pi_{\theta'})~~\textit{s.t.}~~\KL\left(\pi_{\theta},\pi_{\theta'}\right)\leq\delta,~~~\textit{or}~~~
\max_{\theta'} L_{\pi_{\theta}}(\pi_{\theta'})~~\textit{s.t.}~~\D_\gamma\left(\pi_{\theta},\pi_{\theta'})\right)\leq \delta'.
\end{align*}
Furthermore,  we can approximate these constraints up to their second-order Taylor expansion and come up with:
\begin{align*}
\!\!\max_{\theta'} L_{\pi_{\theta}}\!(\pi_{\!\theta'}\!)~\textit{s.t.}~\frac{1}{2}\|\!\left(\theta'\!-\theta\right)^{\!\top}\!\! \!F\!\left(\theta'\!-\theta\right)\!\|_2\!\leq\!\delta,~\textit{or}~
\max_{\theta'} L_{\pi_{\theta}}\!(\pi_{\theta'})~\textit{s.t.}~\frac{1}{2}\|\!\left(\theta'\!-\theta\right)^{\!\top}\!\! F_\gamma\!\left(\theta'-\theta\right)\!\|_2\!\leq\!\delta',\\
\end{align*}
which results in the Alg.~\ref{alg:trpo}. We choose to provided the above mentioned derivation in the expression of the Alg.~\ref{alg:trpo} since the constraints in the optimization can be imposed using the mentioned conjugate gradient techniques.  Alg.~\ref{alg:trpo} is an extension of \TRPO algorithm to the present setting.

We hope that these analyses shed light on how to extend \MDP based methods to the general class of \POMDP{}s, and what are some of the important components crucial for considerations. As mentioned in the introduction, the tools developed in this paper can be used to extend a variety of existing advanced \MDP-based methods to \POMDP{}s. For an example, \thm\ref{thm:DeterministicPolicyGradeintMemory} and~\ref{thm:monotonic} are directly applicable to generalize the constrained policy optimization approaches~\citep{achiam2017constrained} to general episodic environments.


\subsection{Generalized \PPO:}\label{Sec:exp}
The celebrated \PG algorithm \PPO is an extension to \TRPO algorithm which has some favorable properties compared to its predecessor. Broadly speaking, \PPO framework promises 
a more desirable computation and statistical advantages. Following the recipe of extending \TRPO to \PPO, in the following Generalized \PPO (\GPPO), practically more pleasant extension of \GTRPO. 



Usually, in high dimensional but low sample settings, constructing the trust region is hard due to high estimation errors. Many samples are required to have a meaningful construction of the trust region. It is even harder, especially when the region depends on the inverse of the estimated Fisher matrix or when the optimization is constrained with an upper threshold on KL divergence. Therefore, trusting the estimated trust region is questionable. While \TRPO requires concrete construction of the trust region in the parameter space, its final goal is to keep the new policy close to the current policy, i.e., small $\KL\left(\pi_{\theta},\pi_{\theta'}\right)$ or $\D_{\gamma}\left(\pi_{\theta},\pi_{\theta'}\right)$. Proximal Policy Optimization (\PPO) is instead proposed to impose the structure of the trust region directly onto the policy space. This method approximately translates the constraints developed in \TRPO on the parameters space, directly to the policy space, in other mean the action space. \PPO optimization subroutine penalizes the gradients of the objective function when the policy starts to operate beyond the region of trust by setting the gradient to zero. \PPO optimizes for the following objective function,\footnote{In the original \PPO paper, $\delta_U=\delta_L$.}
\begin{align*}
   &\E\Huge[\min\Huge\{\frac{\pi_{\theta'}(a|x)}{\pi_{\theta}(a|x)} A_{\pi_\theta}(a,x), clip(\frac{\pi_{\theta'}(a|x)}{\pi_{\theta}(a|x)};1-\delta_L,1+\delta_U) A_{\pi_\theta}(a,x)\Huge\}\Huge].
\end{align*}
If the advantage function is positive and the importance weight is above $1+\delta_U$ this objective function saturates. When the advantage function is negative and the importance weight is below $1-\delta_L$ this objective function saturates again. In either case, when the objective function saturates, the gradient of this objective function is zero; therefore, further development in that direction is obstructed. This approach, despite its simplicity, approximates the trust region effectively and substantially reduce the computation cost of \TRPO. 

Following the \TRPO, the clipping trick ensures that the importance weight, derived from estimation of $\KL$ does not go beyond a certain limit $|\log\frac{\pi_{\theta'}(a|y)}{\pi_{\theta}(a|y)}|\leq\nu$ when it favors, i.e., 
\begin{align}\label{eq:ppo-clip}
   \!\!\!1-\delta_L:=\exp{(-\nu)}\!\leq\!\frac{\pi_{\theta'}(a|y)}{\pi_{\theta}(a|y)}\!\leq\! 1+\delta_U:=\exp{(\nu)},\!
\end{align}
depending on the sign of the advantage function.
As discussed in the subsection.~\ref{remark:kl}, we propose a principled change in the clipping such that it matches the def of KL divergence in Lem.~\ref{lem:KL-Fisher} and conveys the information in the length of episodes; $|\log\frac{\pi_\theta(a|y)}{\pi_{\theta'}(a|y)}|\leq\frac{\nu}{|\tau|}$; therefore for $\alpha:=\exp{(\nu)}$
\begin{align}\label{eq:clippinglength}
    1-\delta_L:=\alpha^{-1/|\tau|}\leq\frac{\pi_{\theta'}(a|y)}{\pi_{\theta_\theta(a|y)}}\leq1+\delta_U:=\alpha^{1/|\tau|}.
\end{align}
This change ensures more restricted clipping for longer trajectories, while wider for shorter ones. Moreover, as suggested in \thm\ref{thm:monotonic}, and the definition of $\D_\gamma(\pi_{\theta},\pi_{\theta'})$ in Eq.~\ref{eq:Divergence}, we propose an extension in the clipping based on information about the discount factor. Following the prescription $\D_\gamma\left(\pi_{\theta},\pi_{\theta'})\right)\leq \delta'$, for a sample at time step $h$ of an episode, we have $|\log\frac{\pi_{\theta'}(a|y)}{\pi_{\theta}(a|y)}|\leq\frac{\nu}{|\tau|\gamma^{h}}$. Therefore:
\begin{align}\label{eq:clipping}
    &1\!-\!\delta_L^h\!\!:=\!\exp{(\!-\frac{\nu}{|\tau|\gamma^{h}}\!)}\!\leq\!\frac{\pi_{\theta'}(a|y)}{\pi_{\theta}(a|y)}\!\leq\! 1+\delta_U^h\!\!:=\exp{(\frac{\nu}{|\tau|\gamma^{h}})},\nonumber\\
    &\quad\rightarrow \alpha^{-1/|\tau|}\alpha^{-1/\gamma^{h}}\leq\frac{\pi_{\theta'}(a|y)}{\pi_{\theta}(a|y)}\leq \alpha^{1/|\tau|}\alpha^{1/\gamma^{h}}.
\end{align}

\begin{algorithm}[t]
\caption{\GPPO}
\begin{algorithmic}[1]
\STATE Initial $\pi_{\theta}$, $k$, $\lambda$
\STATE Choice of $(\delta_l^h,\delta_u^h)$: $(\delta_L,\delta_U)$, or $(\delta_L^h,\delta_U^h)$,
    \FOR{epoch $=$ 1 until convergence}
        \STATE Deploy policy $\pi_\theta$ and collect experiences
        \STATE Estimate the advantage function $\wh{A}_{\pi_{\theta}}$
        \STATE Construct the empirical estimate $\wh L_{\pi_{\theta}}(\pi_{\theta'},\delta_l^h,\delta_u^h)$ of the following objective function:
\begin{align*}
       &L_{\pi_{\theta}}(\pi_{\theta'},\delta_l^h,\delta_u^h):=\E\Big[\min\Big\{\frac{\pi_{\theta'}(a_h|y_h)}{\pi_{\theta}(a_h|x_h)}A_{\pi_\theta}(y_{h+1},a_h,y_h,y_{h-1},a_{h-1}),\nonumber\\
   &\quad\quad\quad\quad\quad\quad\quad\quad clip(\frac{\pi_{\theta'}(a_h|x_h)}{\pi_{\theta}(a_h|y_h)};1-\delta_l^h,1+\delta_u^h) A_{\pi_\theta}(y_{h+1},a_h,y_h,y_{h-1},a_{h-1})\Big\}\Big]
\end{align*}
        \STATE Initialize $\theta'=\theta$
        \FOR{ $k$ steps}
        \STATE Deploy parameter update $\theta'\leftarrow \theta' + \lambda \nabla_{\theta'}\wh L_{\pi_{\theta}}(\pi_{\theta'})$ or any other ascent algorithm
        \ENDFOR
        \STATE Set $\theta=\theta'$
    \ENDFOR
\end{algorithmic}
\label{alg:gppo}
\end{algorithm}

As it is interpreted, for deeper parts in an episode, we make the clipping softer and allow for larger changes in policy space. This means we are more restricted at the beginning of trajectories compared to the end of trajectories. The choice of $\gamma$ and $\alpha$ are critical here. In practical implementations of \RL algorithms, as also theoretically suggested by \citet{jiang2015dependence,lipton2016combating}, we usually choose discount factors smaller than the one depicted in the problem. Therefore, the discount factor that we use in practice is smaller than the true one, especially when we deploy function approximation.  
Therefore, instead of keeping  $\gamma^{h}$ in Eq.~\ref{eq:clipping}, since the true $\gamma$ in practice is unknown and can be arbitrary close to $1$, we substitute it with a maximum value:
\begin{align}\label{eq:clipping1}
1-\delta_L^h:=&
    \max\lbrace\alpha^{-1/\left(|\tau|\gamma^{h}\right)},1-\beta\rbrace\leq\frac{\pi_{\theta'}(a|y)}{\pi_{\theta}(a|y)}\leq 1+\delta_U^h:=\min\lbrace\alpha^{1/\left(|\tau|\gamma^{h}\right)},1+\beta\rbrace.
\end{align}

The modification proposed in series of equations Eq.~\ref{eq:ppo-clip}, Eq.~\ref{eq:clippinglength}, Eq.~\ref{eq:clipping}, and Eq.~\ref{eq:clipping1} provide insight in the use of trust regions in both \MDP{s} and \POMDP{}s based \PPO. The \GPPO objective for any choice of $\delta_U^h$ and $\delta_L^h$ in \MDP{}s is:
\begin{align}\label{eq:ppo-mdp}
   &\E\Big[\min\Big\{\frac{\pi_{\theta'}(a_h|x_h)}{\pi_{\theta}(a_h|x_h)} A_{\pi_\theta}(a_h,x_h), clip(\frac{\pi_{\theta'}(a_h|x_h)}{\pi_{\theta}(a_h|x_h)};1-\delta_L^h,1+\delta_U^h)A_{\pi_\theta}(a_h,x_h)\Big\}\Big],
\end{align}
while for \POMDP{}s, \GPPO optimizes the following,

\begin{align}\label{eq:ppo-pomdp}
   &L_{\pi_\theta}(\pi_{\theta}):=\E\Big[\min\Big\{\frac{\pi_{\theta'}(a_h|y_h)}{\pi_{\theta}(a_h|x_h)}A_{\pi_\theta}(y_{h+1},a_h,y_h,y_{h-1},a_{h-1}),\nonumber\\
   &\quad\quad\quad\quad\quad\quad\quad\quad clip(\frac{\pi_{\theta'}(a_h|x_h)}{\pi_{\theta}(a_h|y_h)};1-\delta_L^h,1+\delta_U^h) A_{\pi_\theta}(y_{h+1},a_h,y_h,y_{h-1},a_{h-1})\Big\}\Big],
\end{align}

resulting in algorithm \ref{alg:gppo}. In order to make the existing \MDP-based \PPO suitable for \POMDP{}s, we just substitute $A_{\pi_\theta}(a_h,x_h)$ with $A_{\pi_\theta}(y_{h+1},a_h,y_h,y_{h-1},a_{h-1})$ in the codebases. But generally, an extra care is required when turning \MDP based implementation to \POMDP{} based ones,  since for \POMDP{}s, $\nabla_{\theta'}L_{\pi_{\theta}}(\pi_{\theta'})|_{\theta'=\theta}$ might not aligned with $\nabla_{\theta'}\eta(\theta')|_{\theta'=\theta}$ as it is the case for \MDP{}s.

\paragraph{Note:} In this section, we explained how the developed technical tools in this paper could be carried to generalize a variety of \MDP based methods and analysis. For that purpose, we explained how to generalize \TRPO and \PPO to the \POMDP setting. These two extensions are two examples to illustrate the road map of extending \MDP-based methods to \POMDP{}s. 

It is worth noting that the goal if this paper is to provide theoretical tools to study \POMDP{}s, rather than developing new algorithms for partially observable environments. While we show how these tools can be used to generalize a few \MDP-based methods, empirical examination of these generalized methods are not aligned with the study of this paper. We also leave the generalization of the vast literature on \MDP-based methods for later studies.

\section{Conclusion}
In this paper, we studied the algorithmic and convergences properties of \PG methods in the general class of fully and partially observable environments, under both discounted and undiscounted reward. We propose novel notions of value, Q, and advantage functions for \POMDP{}s. We generalize the \MDP{} based \PG theorems to \POMDP{}s. We study the behaviors of polices on the underlying states of \POMDP{}s. We provide convergence analyses of optimizing the surrogate functions using the novel notion of advantage function. We propose a new notation of trust-region in episodic environments and show that the current practice of considering infinite horizon formulation of the trust region, theoretically, is not aligned with the premises of the prior works in episodic environments. To mitigate this shortcoming, we also propose the construction of the discount-factor-dependent trust region. 

We argue that the developed technical tools in this paper to analyze \POMDP{}s are generic tools and can be carried to generalize and extend a variety of existing \MDP based analyses and methods to \POMDP{}s. 
For that purpose, we show how to extend \TRPO and \PPO methods to \POMDP{}s. We propose \GTRPO,  the first trust region policy optimization-based algorithm, which enjoys monotonic improvement guarantees on \POMDP{}s. We further extend our study and propose \GPPO, i.e., an analog to \PPO in \POMDP{}s. These extensions are evident in the generality of the developed tools in this paper.


\section*{Acknowledgements}
K. Azizzadenesheli gratefully acknowledge the financial support of Raytheon, NSF Career Award CCF-1254106 and AFOSR YIP FA9550-15-1-0221. A. Anandkumar is supported in part by Bren endowed chair, DARPA PAIHR00111890035 and LwLL grants, Raytheon, Microsoft, Google, and Adobe faculty fellowships.
\clearpage

\newpage
\appendix
\section{Appendix}
%
%
%


\subsection{Proof of the \thm\ref{thm:PolicyGradeint}}\label{proof:PolicyGradeint}
For a given policy $\pi_\theta$, lets restate the value function at a given state $x_1$: 
\begin{align*}
    \wb V_\theta(x_1) = \int_{a_1,y_1}\pi_\theta(a_1|y_1)O(y_1|x_1)\wb Q_\theta(a_1,x_1)dy_1 da_1
\end{align*}
For this value function, we compute the gradient with respect to parameters $\theta$, i.e., $\Gr \wb V_\theta(x_1)$
\begin{align*}
    \Gr \wb V_\theta(x_1) =& \int_{a_1,y_1}\Gr\pi_\theta(a_1|y_1)O(y_1|x_1)\wb Q_\theta(a_1,x_1)dy_1 da_1 \\
    &\quad+ \int_{a_1,y_1}\pi_\theta(a_1|y_1)O(y_1|x_1)\Gr \wb Q_\theta(a_1,x_1)dy_1 da_1
\end{align*}
To further expand this gradient, consider the definition of $\wb Q_\theta(a_1,x_1)$ using Bellman equation,
\begin{align*}
    \wb Q_\theta(a_1,x_1) = \E[r_1|x_1,a_1] + \gamma \int_{x_2}  \wb V_\theta(x_2)T(x_2|x_1,a_1) dx_2
\end{align*}
where $\E[r_1|x_1,a_1]=\int_{y_1}O(y_1|x_1)\wb{R}(y_1,a_1)dy_1$, resulting in,
\begin{align*}
    \Gr \wb Q_\theta(a_1,x_1) = \gamma \int_{x_2} \Gr \wb V_\theta(x_2) T(x_2|x_1,a_1) dx_2
\end{align*}
since the first term in the definition of $\wb Q_\theta(a_1,x_1)$ does not depend on the parameters $\theta$. Following these steps, we derive $\Gr \wb V_\theta(x_2)$, 
\begin{align*}
    \Gr \wb V_\theta(x_2) =& \int_{a_2,y_2}\Gr\pi_\theta(a_2|y_2)O(y_2|x_2)\wb Q_\theta(a_2,x_2)dy_2 da_2 \\
    &\quad+ \int_{a_2,y_2}\pi_\theta(a_2|y_2)O(y_2|x_2)\Gr \wb Q_\theta(a_2,x_2)dy_2 da_2
\end{align*}
We recursively compute the gradient of the value functions for later time steps and conclude that,
\begin{align*}
    \Gr \wb V_\theta(x_h) &= \int_{a_h,y_h}\Gr\pi_\theta(a_h|y_h)O(y_h|x_h)\wb Q_\theta(a_h,x_h)dy_h da_h \\
    &+ \gamma \int_{a_h,y_h}\pi_\theta(a_h|y_h)O(y_h|x_h) \left(\int_{x_{h+1}} \Gr \wb V_\theta(x_{h+1})T(x_{h+1}|x_h,a_h) dx_{h+1}\right)dy_h da_h
\end{align*}
and repeating this decomposition results in
\begin{align*}
    \Gr \eta(\theta) =& \int_{x_1}P_1(x_1)\Gr \wb V_\theta(x_1)dx_1\\
    =& \int_{a_1,y_1,x_1}P_1(x_1)\Gr\pi_\theta(a_1|y_1)O(y_1|x_1)\wb Q_\theta(a_1,x_1)dy_1 da_1dx_1\\
    &\quad+\gamma \int_{a_1,y_1,x_1}P_1(x_1)
    \pi_\theta(a_1|y_1)O(y_1|x_1) \left(\int_{x_2} \Gr \wb V_\theta(x_2)T(x_2|x_1,a_1) dx_2\right)dy_1 da_1dx_1\\
    =& \int_{\tau}\sum_{h=1}^{|\tau|}\gamma^{h-1} f(\tau_{1..h-1},x_h,y_h;\theta)\Gr\pi_\theta(a_h|y_h) \wb Q_\theta(a_h,x_h)d\tau 
\end{align*}
which is the first statement of the theorem.

For the second statement we have:
\begin{align*}
    \Gr \eta(\theta) =&\int_{\tau}\sum_{h=1}^{|\tau|}\gamma^{h-1} f(\tau_{1..h-1},x_h,y_h;\theta)\Gr\pi_\theta(a_h|y_h) \wb Q_\theta(a_h,x_h)d\tau \\
    =&\int_{\tau}\sum_{h=1}^{|\tau|}\gamma^{h-1} f(\tau_{1..h-1},x_h,y_h;\theta)\pi_\theta(a_h|y_h)\Gr\log\left(\pi_\theta(a_h|y_h)\right) \wb Q_\theta(a_h,x_h)d\tau \\
    =&\int_{\tau}\sum_{h=1}^{|\tau|}\gamma^{h-1} f(\tau_{1..h};\theta)\Gr\log\left(\pi_\theta(a_h|y_h)\right) \wb Q_\theta(a_h,x_h)d\tau 
\end{align*}


\subsection{Proof of the \thm\ref{thm:DeterministicPolicyGradeint}}\label{proof:DeterministicPolicyGradeint}
\begin{proof}
For any time step $h$, let's restate the $\wt Q$, that is the value of committing $\wb a_h$ at state $x_h$, then following the policy induced by $\theta$ afterward. At a time step $h$, state $x_h$,  and $\theta'\in\Theta$, consider $\wb a_h = \mu_{\theta'}(x_h)$, 
\begin{align*}
\wt Q_\theta(\wb a_h,x_h)=&\int_{a_h} \wb Q_\theta(a_h,x_h ) \mu_{\theta'}(a_h|x_h )  da_h=\int_{y_h,a_h} Q_\theta(a_h,x_h ) O(y_h|x_h )\pi_{\theta'}(a_h|y_h )   dy_hda_h
\end{align*}

Given this definition, we have that the value function is $\wb V_\theta(x_1) =\wt Q_\theta(\wb\mu_\theta(x_1),x_1)$, and using Bellman equation we have
\begin{align*}
    \wt Q_\theta(\wb a_h, x_h ) =  \wb r(\wb a_h,x_h )+\gamma \int_{x_{h+1}} \wb T(x_{h+1}|\wb a_h,x_h ) \wb V_\theta(x_{h+1})dx_{h+1}
\end{align*}
for $\wb a_h$ in the set covered by $\mu_\theta(x_h)$ given all $\theta$.

Using this definition, we compute the gradient of $\wt Q_\theta(\wb\mu_\theta(x_h),x_h)$ with respect to parameters $\theta$,

\begin{align*}
    \Gr \wt Q_\theta(\wb\mu_\theta(x_h),x_h)&=\Gr \wb\mu_\theta(x_h)^\top \nabla_{\wb a_h}\wb r(\wb a_h,x_h)|_{\wb a_h =\wb\mu_\theta(x_h)}\\
    &\!\!\!\!\!\!\!\!\!\!\!\!+\gamma \int_{x_{h+1}} \Gr \wb\mu_\theta(x_h)^\top \nabla_{\wb a_h}\wb T(x_{h+1}|\wb a_h,x_h)|_{\wb a_h =\wb\mu_\theta(x_h)} \wt Q_\theta(\wb\mu_\theta(x_{h+1}),x_{h+1})dx_{h+1}\\
    &\!\!\!\!\!\!\!\!\!\!\!\!+\gamma \int_{x_{h+1}} \wb T(x_{h+1}|\wb\mu_\theta(x_h),x_h) \Gr\wt Q_\theta(\wb\mu_\theta(x_{h+1}),x_{h+1})dx_{h+1}\\
    &\!\!\!\!\!\!\!\!\!\!\!\!=\Gr \wb\mu_\theta(x_h) \nabla_{\wb a_h} \Big(\wb  r(\wb a_h,x_h)+\gamma \int_{x_{h+1}} \wb T(x_{h+1}|\wb a_h,x_h) \wt Q_\theta(\wb\mu_\theta(x_{h+1}),x_{h+1})dx_{h+1}\Big)|_{\wb a_h =\wb\mu_\theta(x_h)} \\
    &\!\!\!\!\!\!\!\!\!\!\!\!+\gamma \int_{x_{h+1}} \wb T(x_{h+1}|\wb\mu_\theta(x_h),x_h) \Gr\wt Q_\theta(\wb\mu_\theta(x_{h+1}),x_{h+1})dx_{h+1}\\
    &\!\!\!\!\!\!\!\!\!\!\!\!=\Gr \wb\mu_\theta(x_h)^\top \nabla_{\wb a_h} \wt Q_\theta(\wb a_h,x_h)|_{\wb a_h =\wb\mu_\theta(x_h)}+\gamma \int_{x_{h+1}} \wb T(x_{h+1}|\wb\mu_\theta(x_h),x_h) \Gr\wt Q_\theta(\wb\mu_\theta(x_{h+1}),x_{h+1})dx_{h+1}\\
\end{align*}

therefore similar to the proof of the \thm\ref{thm:PolicyGradeint} we have, 

\begin{align*}
    \Gr \eta(\theta) =& \int_{x_1}P_1(x_1)\Gr \wb V_\theta(x_1)dx_1\\
    =& \int_{x_1}P_1(x_1)\Gr \wb\mu_\theta(x_1)^\top \nabla_{\wb a_1} \wt Q_\theta(\wb a_1,x_1)|_{\wb a_1 =\wb\mu_\theta(x_1)} +\gamma\int_{x_1,x_2}\!\!\!\!\!\!\!\!P_1(x_1) \wb T(x_2|\wb\mu_\theta(x_1),x_1) \Gr\wt Q_\theta(\wb\mu_\theta(x_2),x_2)dx_1dx_2\\
    =&\int_{\tau}\sum_{h=1}^{|\tau|}\gamma^{h-1}f(\tau_{1..h-1},x_h;\theta)\Gr \wb\mu_\theta(x_h)^\top \nabla_{\wb a_h} \wt Q_\theta(\wb a_h,x_h)|_{\wb a_h =\wb\mu_\theta(x_h)}d\tau 
\end{align*}
which is the statement of the theorem.


\end{proof}


\subsection{Proof of the \lem\ref{lem:improvment}}\label{proof:improvment}

In the following we use the fact that for every time step, tuples of observations, and action  $\E_\pi\left[r_h|y_{h+1}, a_{h},y_h \right]=\E\left[r_h|a_{h},y_h \right]$. We utilize this equality and restate the relationship between $V$ and $Q$ in a \POMDP as follows:
\begin{align}\label{eq:Q_V_relation}
Q_\pi&(y_{h+1},a_{h},y_{h} ):=\E_\pi\left[r_h|y_{h+1}, a_{h},y_h \right]
+\gamma V_\pi(y_{h+1},a_{h},y_{h}).
\end{align}
Given the definition of advantage function, we derive,

\begin{align*}
&\E_{\tau\sim\pi_{\theta'}}\sum_{h=1}^{|\tau|}\gamma^{h-1} A_{\pi_{\theta}}(y_{h+1},a_h,y_h,a_{h-1},y_{h-1}) \\
&=\E_{\tau\sim\pi_{\theta'}}\sum_{h=1}^{|\tau|}\gamma^{h-1} \left[Q_{\pi_{\theta}}(y_{h+1},a_h,y_{h})-V_{\pi_{\theta}}(y_{h},a_{h-1},y_{h-1})\right] \\
&=\E_{\tau\sim\pi_{\theta'}}\sum_{h=1}^{|\tau|}\gamma^{h-1} \left[\E_{\pi_{\theta}}\left[r_h|y_{h+1},a_{h},y_h\right]
+\gamma V_{\pi_{\theta}}(y_{h+1},a_{h},y_{h})
-V_{\pi_{\theta}}(y_{h},a_{h-1},y_{h-1})\right] \\
%
%
&=\E_{\tau\sim\pi_{\theta'}}\sum_{h=1}^{|\tau|}\gamma^{h-1} \E\left[r_h|a_{h},y_h\right]
+\E_{\tau\sim\pi_{\theta'}}\sum_{h=1}^{|\tau|}\gamma^{h-1} \left[\gamma V_{\pi_{\theta}}(y_{h+1},a_{h},y_{h})
-V_{\pi_{\theta}}(y_{h},a_{h-1},y_{h-1})\right] \\
&=\E_{\tau\sim\pi_{\theta'}}\sum_{h=1}^{|\tau|}\gamma^{h-1} \E\left[r_h|a_{h},y_h\right]
+\E_{\tau\sim\pi_{\theta'}}\sum_{h=1}^{|\tau|}\gamma^{h-1} \left[\gamma V_{\pi_{\theta}}(y_{h+1},a_{h},y_{h})
-V_{\pi_{\theta}}(y_{h},a_{h-1},y_{h-1})\right] \\
%
&=\E_{\tau\sim\pi_{\theta'}}\sum_{h=1}^{|\tau|}\gamma^{h-1}
\E\left[r_h|a_{h},y_h\right]-\E\left[V_{\pi_{\theta}}(y_0,-,-)\right]\\
&=\E_{\tau\sim\pi_{\theta'}}\sum_{h=1}^{|\tau|}\gamma^{h-1}
\E\left[r_h|a_{h},y_h\right]-\eta(\pi_{\theta})\\
&=\eta(\pi_{\theta'})-\eta(\pi_{\theta})
\end{align*}
which is the statement of the Lemma.


\subsection{Proof of the \thm\ref{thm:DeterministicPolicyGradeintMemory}}\label{proof:DeterministicPolicyGradeintMemory}
As stated in Eq.~\ref{eq:tildA}, for all time steps $h$, observation tuples $y_{h-1},y_h,y_{h+1}$, $a_{h-1}$, $\theta$, and $\theta'$, we have,

\begin{align*}
A_{\pi_{\theta}}(y_{h+1},\check a_h,y_h , a_{h-1},y_{h-1})|_{\check a_h = \pi_{\theta'}(y_h)}
=\int_{a_h} A_{\pi_{\theta}}(y_{h+1},a_h,y_h ,a_{h-1},y_{h-1})\pi_{\theta'}(a_h|y_h) da_h
\end{align*}

Using the definition of $\check A_{\pi_{\theta}}$ and the surrogate objective Eq.~\ref{eq:surrogate}, we state the following:

\begin{align*}
L_{\pi_{\theta}}(\pi_{\theta'})&=\eta(\pi_{\theta})+\E_{\tau\sim\pi_{\theta}}\sum_{h=1}^{|\tau|}\gamma^{h-1}\E_{a_h'\sim \pi_{\theta'}(a_h|y_h)}A_{\pi_{\theta}}(y_{h+1},a_h',y_h ,a_{h-1},y_{h-1})\\
&=\eta(\pi_{\theta})+\E_{\tau\sim\pi_{\theta}}\sum_{h=1}^{|\tau|}\gamma^{h-1}A_{\pi_{\theta}}(y_{h+1},\check a_h,y_h,a_{h-1},y_{h-1})|_{\check a_h = \pi_{\theta'}(y_h)}\\
&=\eta(\pi_{\theta})+\int_{\tau}\sum_{h=1}^{|\tau|} \gamma^{h-1} f(\tau_{1..h},x_{h+1},y_{h+1};\theta) A_{\pi_{\theta}}(y_{h+1},\check a_h,y_h,a_{h-1},y_{h-1})|_{\check a_h = \pi_{\theta'}(y_h)}d\tau
\end{align*}

By taking the gradient of the surrogate objective with respect to $\theta'$, we have:
\begin{align*}
    \nabla_{\theta'}L_{\pi_{\theta}}(\pi_{\theta'})&=\nabla_{\theta'}\int_{\tau}\sum_{h=1}^{|\tau|} \gamma^{h-1} f(\tau_{1..h},x_{h+1},y_{h+1};\theta) A_{\pi_{\theta}}(y_{h+1},\check a_h,y_h,a_{h-1},y_{h-1})|_{\check a_h = \pi_{\theta'}(y_h)}d\tau\\
    &=\int_{\tau}\sum_{h=1}^{|\tau|} \gamma^{h-1} f(\tau_{1..h},x_h,y_h;\theta)\nabla_{\theta'} \pi_{\theta'}(y_h)^\top \nabla_{\check a_h} A_{\pi_{\theta}}(y_{h+1},\check a_h,y_h,a_{h-1},y_{h-1})|_{\check a_h = \pi_{\theta'}(y_h)}d\tau
\end{align*}

which results in the statement of the theorem by evaluating $\nabla_{\theta'}L_{\pi_{\theta}}(\pi_{\theta'})$ at $\theta=\theta'$.





\subsection{Proof of the \thm\ref{thm:Convergence}}\label{proof:Convergence}

For a parameter $\theta$, its corresponding $\varpi(\theta)$, and an $\alpha$, consider the surrogate objective:

\begin{align*}
L_{\pi_{\theta}}(\pi_{\theta+\alpha \varpi(\theta)})&=\eta(\pi_{\theta})+\int_{\tau}\sum_{h=1}^{|\tau|} \gamma^{h-1} f(\tau_{1..h},x_{h+1},y_{h+1};\theta) A_{\pi_{\theta}}(y_{h+1},\check a_h,y_h,a_{h-1},y_{h-1})|_{\check a_h =\pi_{\theta+\alpha \varpi(\theta)}(y_h)}d\tau
\end{align*}

Therefore for the directional gradient we have:
\begin{align*}
    \nabla_{\alpha}L_{\pi_{\theta}}(\pi_{\theta+\alpha \varpi(\theta)})&=\nabla_{\alpha}\int_{\tau}\sum_{h=1}^{|\tau|} \gamma^{h-1} f(\tau_{1..h},x_{h+1},y_{h+1};\theta) A_{\pi_{\theta}}(y_{h+1},\check a_h,y_h,a_{h-1},y_{h-1})|_{\check a_h =\pi_{\theta+\alpha \varpi(\theta)}(y_h)}d\tau\\
    &\!\!\!\!\!\!\!\!\!\!\!\!\!\!\!\!\!\!\!\!\!\!\!\!=\int_{\tau}\sum_{h=1}^{|\tau|} \gamma^{h-1} f(\tau_{1..h},x_{h+1},y_{h+1};\theta)\nabla_{\alpha} \pi_{\theta+\alpha \varpi(\theta)}(y_h)^\top\nabla_{\check a_h} A_{\pi_{\theta}}(y_{h+1},\check a_h,y_h,a_{h-1},y_{h-1})|_{\check a_h =\pi_{\theta+\alpha \varpi(\theta)}(y_h)}d\tau
\end{align*}

Now if we evaluate it at $\alpha=0$ we have

\begin{align*}
    &\nabla_{\alpha}L_{\pi_{\theta}}(\pi_{\theta+\alpha \varpi(\theta)})|_{\alpha=0}\\
    &\quad=\int_{\tau}\sum_{h=1}^{|\tau|} \gamma^{h-1} f(\tau_{1..h},x_{h+1},y_{h+1};\theta)\nabla_{\alpha} \pi_{\theta+\alpha \varpi(\theta)}|_{\alpha=0}(y_h)^\top\nabla_{\check a_h} A_{\pi_{\theta}}(y_{h+1},\check a_h,y_h,a_{h-1},y_{h-1})|_{\check a_h = \pi_{\theta}(y_h)}d\tau\\
    &\quad\stackrel{1}{=}\int_{\tau}\sum_{h=1}^{|\tau|} \gamma^{h-1} f(\tau_{1..h},x_{h+1},y_{h+1};\theta) \left(\pi_{\theta^\star(\theta)}(y_h)-\pi_\theta(y_h)\right)^\top\nabla_{\check a_h} A_{\pi_{\theta}}(y_{h+1},\check a_h,y_h,a_{h-1},y_{h-1})|_{\check a_h = \pi_{\theta}(y_h)}d\tau\\
    &\quad\stackrel{2}{\geq} \int_{\tau}\sum_{h=1}^{|\tau|} \gamma^{h-1} f(\tau_{1..h},x_{h+1},y_{h+1};\theta) \left(A_{\pi_{\theta}}(y_{h+1},\pi_{\theta^\star(\theta)}(y_h),y_h,a_{h-1},y_{h-1})-A_{\pi_{\theta}}(y_{h+1},\pi_{\theta}(y_h),y_h,a_{h-1},y_{h-1})\right)d\tau\\
    &=\T A_{\pi_{\theta}}
\end{align*}
where the equality (1) follows from the Asm.~\ref{asm:existence}, and the inequality in (2) follows from the concavity Asm.~\ref{asm:convex}.
As a consequence we have:
\begin{align*}
    \T A_{\pi_{\theta}}\leq \nabla_{\alpha}L_{\pi_{\theta}}(\pi_{\theta+\alpha u})|_{\alpha=0}
\end{align*}

\subsection{Proof of the \thm\ref{thm:Convergence_dist}}\label{proof:Convergence_dist}

The results in the \lem\ref{lem:improvment} indicates that
\begin{align*}
\eta(\pi_{\theta'})-\eta(\pi_{\theta}) &= \E_{\tau\sim\pi_{\theta'}}\sum_{h=1}^{|\tau|}\gamma^{h-1} A_{\pi_{\theta}}(y_{h+1},a_h,y_h ,a_{h-1},y_{h-1})   \\
&=\int_{\tau}\sum_{h=1}^{|\tau|} \gamma^{h-1} f(\tau_{1..h},x_{h+1},y_{h+1};\theta')A_{\pi_{\theta}}(y_{h+1},a_h,y_h ,a_{h-1},y_{h-1}) d\tau
\end{align*}
In the following we use the definition of distribution mismatch coefficient to derive the statement of the theorem. For any optimal parameter $\theta^*$ and applying \lem\ref{lem:improvment} we have,
\begin{align}\label{eq:supAdvant}
&\eta(\pi_{\theta^*})-\eta(\pi_{\theta}) \nonumber \\
&=\int_{\tau}\sum_{h=1}^{|\tau|} \gamma^{h-1} f(\tau_{1..h-1},x_{h},y_{h};\theta^*)\pi_{\theta^*}(a_h|y_h) T(x_{h+1}|x_h,a_h)O(y_{t+1}|x_{t+1})A_{\pi_{\theta}}(y_{h+1},a_h,y_h ,a_{h-1},y_{h-1}) d\tau\nonumber\\
&\leq\max_{\theta''\in\Theta}\int_{\tau}\sum_{h=1}^{|\tau|} \gamma^{h-1} f(\tau_{1..h-1},x_{h},y_{h};\theta^*)\pi_{\theta''}(a_h|y_h) T(x_{h+1}|x_h,a_h)O(y_{t+1}|x_{t+1})A_{\pi_{\theta}}(y_{h+1},a_h,y_h ,a_{h-1},y_{h-1}) d\tau.
\end{align}

Let $\theta^+$, as defined in Eq.~\ref{eq:policyplus0}, denote a parameter in the set of parameters that achieve the maximum in Eq.~\ref{eq:supAdvant}, i.e.,
\begin{align}\label{eq:policyplus}
    \theta^+\in \argmax_{\theta''\in\Theta}\int_{\tau}\sum_{h=1}^{|\tau|} \gamma^{h-1} f(\tau_{1..h-1},x_{h},y_{h};\theta^*)\pi_{\theta''}(a_h|y_h) T(x_{h+1}|x_h,a_h)O(y_{t+1}|x_{t+1})A_{\pi_{\theta}}(y_{h+1},a_h,y_h ,a_{h-1},y_{h-1}) d\tau.
\end{align} 

Given the definition of $\pi^+_{\theta^+,\theta}$ in Eq.~\ref{eq:piplus}, we have,
\begin{align}\label{eq:supAdvant1}
&\eta(\pi_{\theta^*})-\eta(\pi_{\theta}) \nonumber \\
&\leq\int_{\tau}\sum_{h=1}^{|\tau|} \gamma^{h-1} f(\tau_{1..h-1},x_{h},y_{h};\theta^*)\pi_{\theta^+}(a_h|y_h) T(x_{h+1}|x_h,a_h)O(y_{h+1}|x_{h+1})A_{\pi_{\theta}}(y_{h+1},a_h,y_h ,a_{h-1},y_{h-1}) d\tau\nonumber\\
&\leq\int_{\tau}\sum_{h=1}^{|\tau|} \gamma^{h-1} \frac{f(\tau_{1..h-1},x_{h};\theta^*)}{f(\tau_{1..h-1},x_{h};\theta)}f(\tau_{1..h-1},x_{h},y_{h};\theta)\pi^+_{\theta^+,\theta}(a_h,y_h,x_h,a_{h-1},y_{h-1}) \nonumber\\
&\quad\quad\quad\quad\quad\quad\quad\quad\quad\quad\quad\quad T(x_{h+1}|x_h,a_h)O(y_{h+1}|x_{h+1})A_{\pi_{\theta}}(y_{h+1},a_h,y_h ,a_{h-1},y_{h-1}) d\tau\nonumber\\
&\leq\|\frac{f(\tau:\theta^*)}{f(\tau:\theta)}\|_\infty\int_{\tau}\sum_{h=1}^{|\tau|} \gamma^{h-1} f(\tau_{1..h-1},x_{h},y_{h};\theta)\pi^+_{\theta^+,\theta}(a_h,y_h,x_h,a_{h-1},y_{h-1}) \nonumber\\
    &\quad\quad\quad\quad\quad\quad\quad\quad\quad\quad\quad\quad T(x_{h+1}|x_h,a_h)O(y_{h+1}|x_{h+1})A_{\pi_{\theta}}(y_{h+1},a_h,y_h ,a_{h-1},y_{h-1}) d\tau,
\end{align}

where the last inequality step results from the fact that the integrand, i.e., the random variable on the right hand side of the Eq.~\ref{eq:supAdvant1} is positive. 
%
%
Using the \lem\ref{lem:improvment} again, we restate that
\begin{align*}
&\int_{\tau}\sum_{h=1}^{|\tau|} \gamma^{h-1} f(\tau_{1..h-1},x_{h},y_{h};\theta)\pi_{\theta}(a_h|y_h) T(x_{h+1}|x_h,a_h)O(y_{h+1}|x_{h+1})A_{\pi_{\theta}}(y_{h+1},a_h,y_h ,a_{h-1},y_{h-1}) d\tau\\
&\quad\quad\quad\quad\quad\quad=\int_{\tau}\sum_{h=1}^{|\tau|} \gamma^{h-1} f(\tau_{1..h},x_{h+1},y_{h+1};\theta)
A_{\pi_{\theta}}(y_{h+1},a_h,y_h ,a_{h-1},y_{h-1})  d\tau \\
&\quad\quad\quad\quad\quad\quad= \eta(\theta)-\eta(\theta) = 0,
\end{align*}
therefore, we can subtract this term from the write hand side of Eq.~\ref{eq:supAdvant1} without changing the direction of the inequality. As a result of this subtraction we have

\begin{align}\label{eq:supAdvant3}
&\eta(\pi_{\theta^*})-\eta(\pi_{\theta})\leq\|\frac{f(\tau:\theta^*)}{f(\tau:\theta)}\|_\infty\Delta\eta^+(\theta^+,\theta)\\
&\textit{where~}\Delta\eta^+(\theta^+,\theta):=\int_{\tau}\sum_{h=1}^{|\tau|} \gamma^{h-1} f(\tau_{1..h-1},x_{h},y_{h};\theta)\left(\pi^+_{\theta^+,\theta}(a_h,y_h,x_h,a_{h-1},y_{h-1})-\pi_{\theta}(a_h|y_h) \right)\nonumber\\
&\quad\quad\quad\quad\quad\quad\quad\quad\quad\quad\quad\quad T(x_{h+1}|x_h,a_h)O(y_{h+1}|x_{h+1})A_{\pi_{\theta}}(y_{h+1},a_h,y_h ,a_{h-1},y_{h-1}) d\tau,\nonumber
\end{align}
as noted in Eq.~\ref{eq:supAdvant30}. Let us restate that in the case of \MDP, and when the policy class is rich enough (or the \MDP is simple enough, e.g., fairly small tabular \MDP), $\Delta\eta^+(\theta^+,\theta)$ vanishes.



Using the definitions of the Bellman policy error compatibility vector in Def.~\ref{def:BPE}, we have:
\begin{align}\label{eq:supAdvant4}
&\eta(\pi_{\theta^*})-\eta(\pi_{\theta}) \nonumber \\
&\leq\|\frac{f(\tau:\theta^*)}{f(\tau:\theta)}\|_\infty\int_{\tau}\sum_{h=1}^{|\tau|} \gamma^{h-1} f(\tau_{1..h-1},x_{h},y_{h};\theta)\left(\pi^+_{\theta^+,\theta}(a_h,y_h,x_h,a_{h-1},y_{h-1})-\pi_{\theta}(a_h|y_h)\right)\nonumber\\
&\quad\quad\quad\quad\quad\quad\quad\quad\quad\quad\quad\quad T(x_{h+1}|x_h,a_h)O(y_{h+1}|x_{h+1})A_{\pi_{\theta}}(y_{h+1},a_h,y_h ,a_{h-1},y_{h-1}) d\tau\nonumber\\
&-\|\frac{f(\tau:\theta^*)}{f(\tau:\theta)}\|_\infty\int_{\tau}\sum_{h=1}^{|\tau|} \gamma^{h-1} f(\tau_{1..h-1},x_{h+1},y_{h+1};\theta){\omega^+}^\top\big(\!\!\int_{a'_h}\!\!\nabla_\theta\pi_{\theta}(a_h'|y_h) A_{\pi_{\theta}}(y_{h+1},a_h',y_h ,a_{h-1},y_{h-1})da'_h\big) d\tau\nonumber\\
&+\|\frac{f(\tau:\theta^*)}{f(\tau:\theta)}\|_\infty\int_{\tau}\sum_{h=1}^{|\tau|} \gamma^{h-1} f(\tau_{1..h-1},x_{h+1},y_{h+1};\theta){\omega^+}^\top\big(\!\!\int_{a'_h}\!\!\nabla_\theta\pi_{\theta}(a_h'|y_h) A_{\pi_{\theta}}(y_{h+1},a_h',y_h ,a_{h-1},y_{h-1})da'_h\big) d\tau\nonumber\\
&\leq\epsilon^+_{\theta}\|\frac{f(\tau:\theta^*)}{f(\tau:\theta)}\|_\infty+\|\frac{f(\tau:\theta^*)}{f(\tau:\theta)}\|_\infty\nonumber\\
&\quad\quad\quad\quad\quad
\int_{\tau}\sum_{h=1}^{|\tau|} \gamma^{h-1} f(\tau_{1..h-1},x_{h+1},y_{h+1};\theta){\omega^+}^\top\big(\!\!\int_{a'_h}\!\!\nabla_\theta\pi_{\theta}(a_h'|y_h) A_{\pi_{\theta}}(y_{h+1},a_h',y_h ,a_{h-1},y_{h-1})da'_h\big) d\tau\nonumber\\
&\leq\epsilon^+_{\theta}\|\frac{f(\tau:\theta^*)}{f(\tau:\theta)}\|_\infty+\|\frac{f(\tau:\theta^*)}{f(\tau:\theta)}\|_\infty{\omega^+}^\top\nonumber\\
&\quad\quad\quad\quad\quad
\int_{\tau}\sum_{h=1}^{|\tau|} \gamma^{h-1} f(\tau_{1..h-1},x_{h+1},y_{h+1};\theta)\big(\!\!\int_{a'_h}\!\!\nabla_\theta\pi_{\theta}(a_h'|y_h) A_{\pi_{\theta}}(y_{h+1},a_h',y_h ,a_{h-1},y_{h-1})da'_h\big) d\tau,\nonumber\\
\end{align}
where using the result in Eq.~\ref{eq:PolicyGradeintSurrogate} conclude the proof.

\subsection{Proof of Lemma~\ref{lem:KL-Fisher}}\label{proof:KL-Fisher}
\begin{proof}
The proof is based on a few following steps. Under the considerations of Lebesgue’s dominated convergence theorem,
\begin{align}
\nabla^2_{\theta'}KL(\theta,\theta')|_{\theta'=\theta} :=&-\nabla^2_{\theta'}\int_{\tau} f(\tau;\theta) \left[\log\left(f(\tau;\theta')\right)-\log\left(f(\tau;\theta)\right)\right] d\tau|_{\theta'=\theta} \nonumber\\
=& -\int_{\tau} f(\tau;\theta) \nabla^2_{\theta'}\log\left(f(\tau;\theta')\right)d\tau|_{\theta'=\theta}\nonumber\\
=& -\int_{\tau} f(\tau;\theta) \nabla_{\theta'}\left[\frac{1}{f(\tau;\theta')}\nabla_{\theta'}f(\tau;\theta')\right]d\tau|_{\theta'=\theta}\nonumber\\
=& \int_{\tau} f(\tau;\theta) \left[\frac{1}{f(\tau;\theta')^2}\nabla_{\theta'}f(\tau;\theta')\nabla_{\theta'}f(\tau;\theta')^\top\right]d\tau|_{\theta'=\theta}\nonumber\\
&\quad\quad\quad\quad\quad\quad\quad\quad\quad\quad\quad\quad\quad-\int_{\tau} f(\tau;\theta) \left[\frac{1}{f(\tau;\theta')}\nabla_{\theta'}^2f(\tau;\theta')\right]d\tau|_{\theta'=\theta}\nonumber\\
=& \int_{\tau} f(\tau;\theta) \left[\frac{1}{f(\tau;\theta')^2}\nabla_{\theta'}f(\tau;\theta')\nabla_{\theta'}f(\tau;\theta')^\top\right]d\tau|_{\theta'=\theta}-\nabla_{\theta'}^2\int_{\tau}f(\tau;\theta')d\tau|_{\theta'=\theta}\nonumber\\
=&F(\theta),
\end{align}
which concludes the proof.
\end{proof}
\subsection{Proof of the \thm\ref{thm:monotonic}}\label{proof:monotonic}
\begin{proof} 
Following the result in the \lem\ref{lem:improvment} we have
%
%
%
%
\begin{align*}
\eta(\pi_{\theta'})=\eta(\pi_{\theta})+\E_{\tau\sim\pi_{\theta'}}\sum_{h=1}^{|\tau|}\gamma^{h-1}A_{\pi_{\theta}}(y_{h+1},a_h,y_{h},y_{h-1},a_{h-1}) 
\end{align*}
therefore, using the definition of the surrogate function, we have,

\begin{align*}
\eta(\pi_{\theta'})- L_{\pi}(\pi_{\theta'})&=\int_{\tau}f(\tau;\pi_{\theta'})\sum_{h=1}^{|\tau|}\gamma^{h-1}{A}_{\pi_{\theta}}(y_{h+1},a_{h},y_{h},y_{h-1},a_{h-1})d\tau\\
&\quad\quad\quad\quad-\int_{\tau}f(\tau;\pi_{\theta})\sum_{h=1}^{|\tau|}\gamma^{h-1}{A}_{\pi_{\theta}}(y_{h+1},\pi_{\theta'}(y_{h}),y_{h},y_{h-1},a_{h-1})d\tau\\
&=\int_{\tau}f(\tau;\pi_{\theta'})\sum_{h=1}^{|\tau|}\gamma^{h-1}{A}_{\pi_{\theta}}(y_{h+1},a_{h},y_{h},y_{h-1},a_{h-1})d\tau\\
&\quad\quad\quad\quad-\int_{\tau}f(\tau;\pi_{\theta'})\sum_{h=1}^{|\tau|}\gamma^{h-1}{A}_{\pi_{\theta}}(y_{h+1},\pi_{\theta'}(y_{h}),y_{h},y_{h-1},a_{h-1})d\tau\\
&\quad\quad\quad\quad+\int_{\tau}f(\tau;\pi_{\theta'})\sum_{h=1}^{|\tau|}\gamma^{h-1}{A}_{\pi_{\theta}}(y_{h+1},\pi_{\theta'}(y_{h}),y_{h},y_{h-1},a_{h-1})d\tau\\
&\quad\quad\quad\quad-\int_{\tau}f(\tau;\pi_{\theta})\sum_{h=1}^{|\tau|}\gamma^{h-1}{A}_{\pi_{\theta}}(y_{h+1},\pi_{\theta'}(y_{h}),y_{h},y_{h-1},a_{h-1})d\tau\\
\end{align*}

Define advantage gap, $\wb\epsilon$, a notion of gap between \MDP{} and \POMDP{}s incorporating the effect of future observations in the advantage functions,

\begin{align}\label{eq:epsilon_bar}
   \wb\epsilon :=&\max_{(\theta,\theta')\in\Theta\times\Theta}\Big|\int_{\tau}f(\tau;\pi_{\theta'})\sum_{h=1}^{|\tau|}\gamma^{h-1}\Big({A}_{\pi_{\theta}}(y_{h+1},a_{h},y_{h},y_{h-1},a_{h-1})-{A}_{\pi_{\theta}}(y_{h+1},\pi_{\theta'}(y_{h}),y_{h},y_{h-1},a_{h-1} \Big)d\tau\Big|.
\end{align}
Note that advantage gap $\wb\epsilon$ is equal to zero in \MDP{}s. Using the definition of $\wb\epsilon$,  we have,

\begin{align}\label{eq:adv_gap}
\Big|\eta(\pi_{\theta'})- L_{\pi}(\pi_{\theta'})\Big|&\leq
\wb\epsilon+\Big|\int_{\tau}f(\tau;\pi_{\theta'})\sum_{h=1}^{|\tau|}\gamma^{h-1}{A}_{\pi_{\theta}}(y_{h+1},\pi_{\theta'}(y_{h}),y_{h},y_{h-1},a_{h-1})d\tau\nonumber\\
&\quad\quad\quad\quad-\int_{\tau}f(\tau;\pi_{\theta})\sum_{h=1}^{|\tau|}\gamma^{h-1}{A}_{\pi_{\theta}}(y_{h+1},\pi_{\theta'}(y_{h}),y_{h},y_{h-1},a_{h-1})d\tau\Big|.
\end{align}

Note that, when we adopt $f(\tau;\theta)d\tau:=d\PP_\theta$, and $\PP_\theta(\Gamma)=\int_\Gamma f(\tau;\theta)d\tau$, we encoded the absolute continuity of measure $\PP_\theta$ for all $\theta\in\Theta$ with respect to a positive measure, e.g., Lebesgue's. Using this consideration, we have 
\begin{align*}
\Big|\eta(\pi_{\theta'})- L_{\pi}(\pi_{\theta'})\Big|&\leq
\wb\epsilon+\int_{\tau}\Big|f(\tau;\pi_{\theta'})-f(\tau;\pi_{\theta})\Big|d\tau\\
&=\wb\epsilon+\epsilon TV\left(\tau\sim f(\cdot;\pi_{\theta'}),\tau\sim f(\tau;\pi_{\theta})\right)\\
\end{align*}
and deploying the Pinsker's inequality we have have the first statement of the theorem, 
\begin{align*}
|\eta(\pi_{\theta'})- L_{\pi_{\theta}}(\pi_{\theta'})|
&\leq\wb\epsilon+\epsilon TV\left(\tau\sim f(\cdot;\pi_{\theta'}),\tau\sim f(\tau;\pi_{\theta})\right)\\
&\leq \wb\epsilon+\epsilon \sqrt{\frac{1}{2}\KL\left(\tau\sim f(\cdot;\pi_{\theta'}),\tau\sim f(\tau;\pi_{\theta})\right)}\\
\end{align*}


For the second statement of the  theorem, let us restate the decomposition in Eq.~\ref{eq:adv_gap} under the consideration in Fubini's theorem,
\begin{align*}
\Big|\eta(\pi_{\theta'})- L_{\pi}(\pi_{\theta'})\Big|&\leq
\wb\epsilon+\Big|\sum_{h\geq1}\gamma^{h-1}\int_{\tau}f(\tau;\pi_{\theta'}){A}_{\pi_{\theta}}(y_{h+1},\pi_{\theta'}(y_{h}),y_{h},y_{h-1},a_{h-1})d\tau\\
&\quad\quad\quad\quad-\sum_{h\geq1}\gamma^{h-1}\int_{\tau}f(\tau;\pi_{\theta}){A}_{\pi_{\theta}}(y_{h+1},\pi_{\theta'}(y_{h}),y_{h},y_{h-1},a_{h-1})d\tau\Big|\\
&=
\wb\epsilon+\Big|\sum_{h\geq1}\gamma^{h-1}\int_{\tau}(f(\tau;\pi_{\theta'})-f(\tau;\pi_{\theta})){A}_{\pi_{\theta}}(y_{h+1},\pi_{\theta'}(y_{h}),y_{h},y_{h-1},a_{h-1})d\tau\Big|\\
&\leq
\wb\epsilon+\sum_{h\geq1}\gamma^{h-1}\int_{\tau}\Big|f(\tau;\pi_{\theta'})-f(\tau;\pi_{\theta})\Big|\Big|{A}_{\pi_{\theta}}(y_{h+1},\pi_{\theta'}(y_{h}),y_{h},y_{h-1},a_{h-1})\Big|d\tau
\end{align*}
Applying the similar argument we used in proving the first statement, we have,

\begin{align*}
\Big|\eta(\pi_{\theta'})- L_{\pi}(\pi_{\theta'})\Big|
&\leq
\wb\epsilon+\epsilon'\sum_{h\geq1}\gamma^{h-1}\int_{\tau}\Big|f(\tau;\pi_{\theta'})-f(\tau;\pi_{\theta})\Big|d\tau\\
&\leq
\wb\epsilon+\epsilon'\sum_{h\geq1}\gamma^{h-1}TV\left(\tau_{1..h}\sim f(\cdot;\pi_{\theta'}),\tau_{1..h}\sim f(\cdot;\pi_{\theta})\right)\\
&\leq
\wb\epsilon+\epsilon'\sum_{h\geq1}\gamma^{h-1}\sqrt{\frac{1}{2}\KL\left(\tau_{1..h}\sim f(\cdot;\pi_{\theta'}),\tau_{1..h}\sim f(\cdot;\pi)\right)}\\
\end{align*}

Deploying the definition of $\D_\gamma(\pi_{\theta},\pi_{\theta'})$,

\begin{align*}
\Big|\eta(\pi_{\theta'})- L_{\pi}(\pi_{\theta'})\Big|\leq
\wb\epsilon+\epsilon'\sum_{h\geq1}\gamma^{h-1}\sqrt{\frac{1}{2}\KL\left(\tau_{1..h}\sim f(\cdot;\pi_{\theta'}),\tau_{1..h}\sim f(\cdot;\pi)\right)}= \wb\epsilon+\epsilon'  \sqrt{\D_\gamma(\pi_{\theta},\pi_{\theta'})}
\end{align*}

and the second part of the theorem goes through.
\end{proof}

\newpage

\bibliography{ref}
\bibliographystyle{plainnat}
\end{document}